%% file: main.tex
\documentclass[]{fairmeta}
\input{math_commands.tex}

\usepackage{fontawesome}
\usepackage{wrapfig}
\usepackage{fancyhdr}

\definecolor{softgreen}{RGB}{110, 160, 120}

\newtcolorbox{takeawaybox_basemodel}[1]{
    colback=orange!5!white,    % 背景颜色（浅橙色/米色）
    colframe=black,            % 边框颜色
    arc=5pt,                   % 圆角半径
    outer arc=5pt,
    boxrule=0.8pt,             % 边框粗细
    left=5pt,                 % 左内边距
    right=5pt,                % 右内边距
    top=4pt,                   % 上内边距
    bottom=4pt,                % 下内边距
    fontupper=\small,          % 正文字体大小
    enhanced,
    before upper={\textbf{#1: }} 
}

\newtcolorbox{takeawaybox_rlmodel}[1]{
    colback=blue!5!white,    % 背景颜色（浅橙色/米色）
    colframe=black,            % 边框颜色
    arc=5pt,                   % 圆角半径
    outer arc=5pt,
    boxrule=0.8pt,             % 边框粗细
    left=5pt,                 % 左内边距
    right=5pt,                % 右内边距
    top=4pt,                   % 上内边距
    bottom=4pt,                % 下内边距
    fontupper=\small,          % 正文字体大小
    enhanced,
    before upper={\textbf{#1: }} 
}
\definecolor{earlyblue}{HTML}{88A2F1}
\definecolor{midgrey}{HTML}{fadcb4}
\definecolor{latered}{HTML}{EE9C88}
\definecolor{highlightgreen}{HTML}{80c66d}
\definecolor{highlightpurple}{HTML}{9b6d97}

\usepackage{arydshln} % load AFTER colortbl/xcolor
\usepackage{xstring}
\newcommand{\deltaval}[1]{%
  \IfBeginWith{#1}{+}{%
    {\textcolor{highlightgreen}{\textit{(#1)}}}%
  }{%
    \IfBeginWith{#1}{-}{%
      {\textcolor{highlightpurple}{\textit{(#1)}}}%
    }{%
      {\textit{(#1)}}%
    }%
  }%
}
\usepackage{pifont}

\usepackage[export]{adjustbox}

\newtcolorbox{promptbox}{
    colback=gray!8,
    colframe=black!20,
    boxrule=0.5pt,
    arc=3pt,
    left=6pt,
    right=6pt,
    top=6pt,
    bottom=6pt
}

\usepackage[most]{tcolorbox}

\newtcolorbox{modelquote}[1][]{
    colback=gray!5,      % 浅灰色的背景
    colframe=gray!40,    % 边框颜色
    arc=2pt,             % 边角弧度
    boxrule=0.5pt,       % 边框粗细
    left=6pt, right=6pt, top=4pt, bottom=4pt,
    fonttitle=\bfseries\small,
    coltitle=black,
    breakable            % 允许跨页
}

\title{When AI Navigates the Fog of War}

\author[1,2,*]{Ming Li}
\author[2,*]{Xirui Li}
\author[1]{Tianyi Zhou}

\renewcommand\affiliation[2][]{%
  \addtolist[#1]{#2}{\affiliationlist}{\affiliationformat}{\\}%
}

\affiliation[1]{Mohamed bin Zayed University of Artificial Intelligence, Abu Dhabi, United Arab Emirates}
\affiliation[2]{University of Maryland, College Park, United States}

\contribution[*]{Co-first Author}
\abstract{
\input{tex/abstract}
}

\date{\today}
\authoremails{\email{minglii@umd.edu} \&~\email{Ming.Li@mbzuai.ac.ae}, \email{xiruili@umd.edu}, \email{Tianyi.Zhou@mbzuai.ac.ae}}

\metadata[Project Page]{\url{www.war-forecast-arena.com}\\}

\begin{document}

\fancypagestyle{firstpage}{
  \fancyhf{}              % 清空默认 header/footer
  \fancyhead[l]{\small\bfseries 
  \textcolor{red}{\faExclamationTriangle ~\texttt{Disclaimer: This work presents no political stance or endorsement of any kind. Our objective is to analyze the reasoning behaviors of LLMs. All event nodes and entity names appearing in this work are sampled based on publicly available Wikipedia and news pages and are used purely for research purposes.}} 
  % \looseness-1
  } % 你的 disclaimer
  \renewcommand{\headrulewidth}{0pt} % 去掉页眉横线
}

\maketitle
\thispagestyle{firstpage}

\begin{quote}
  ``War is the realm of uncertainty; three quarters of the factors on which action in war is based are wrapped in a fog of greater or lesser uncertainty.'' \\
  \end{quote}
  \begin{flushright}
  \begin{minipage}{0.25\textwidth}
  --- \textit{Carl von Clausewitz}
  \end{minipage}
  \end{flushright}

\begin{quote}
  ``Peace cannot be kept by force; it can only be achieved by understanding.'' \\
\end{quote}
\begin{flushright}
\begin{minipage}{0.25\textwidth}
--- \textit{Albert Einstein}
\end{minipage}
\end{flushright}

\input{tex/introduction}
%\clearpage
\input{tex/background}
%\clearpage
\input{tex/method}
%\clearpage
\input{tex/main_part}

\input{tex/discussion}
\input{tex/quantitive}
\input{tex/conclusion}

\clearpage
\newpage
\bibliographystyle{assets/plainnat}
\bibliography{main}

\clearpage
\newpage
\beginappendix
\input{tex/appendix_experiment_settings.tex}

\input{tex/appendix_complete_examples.tex}

\input{tex/appendix_detailed_performance_results.tex}

\end{document}

%% file: math_commands.tex
\usepackage{amsmath,amsfonts,bm}
\usepackage{xspace}
\usepackage{hyperref}
\usepackage{url}
\usepackage{subcaption}

\usepackage{colortbl}
\usepackage{xcolor}

\usepackage{graphicx}
\usepackage{algorithmic}
\usepackage{algorithm}
\usepackage{float}

\usepackage{booktabs}

\usepackage{enumitem}

\usepackage{amsthm}

\theoremstyle{plain}

\theoremstyle{definition}

\theoremstyle{remark}

\def\eqref#1{equation~\ref{#1}}
\def\1{\bm{1}}

\DeclareMathAlphabet{\mathsfit}{\encodingdefault}{\sfdefault}{m}{sl}
\SetMathAlphabet{\mathsfit}{bold}{\encodingdefault}{\sfdefault}{bx}{n}

%% file: tex/abstract.tex
\textit{Can AI reason about and forecast the trajectory of an ongoing war before it transitions into history?} 
Analyzing this capability is difficult because retrospective geopolitical prediction is heavily confounded by \textit{training-data leakage}. We address this challenge through a temporally grounded case study of the early stages of the 2026 Middle East conflict, which unfolded after the training cutoff of current frontier models. 
We construct 11 critical temporal nodes, 42 node-specific verifiable questions, and 5 general exploratory questions, requiring models to reason only from information that would have been publicly available at each moment. 
This design substantially mitigates training-data leakage concerns, creating a setting well-suited for studying how models analyze an unfolding crisis under the fog of war, and provides, to our knowledge, the first \textbf{temporally grounded analysis of LLM reasoning in an ongoing geopolitical conflict}.
Our analysis reveals three main findings. 
First, current state-of-the-art large language models often show \textbf{strong strategic reasoning} in this setting, moving beyond surface political rhetoric to attend to underlying incentives, deterrence pressures, and material constraints. 
Second, this capability is uneven across domains: models are more reliable in economically and logistically structured settings than in politically ambiguous multi-actor environments. 
Finally, model narratives evolve over time, shifting from early expectations of rapid containment toward more systemic accounts of regional entrenchment and attritional de-escalation. 
\textit{Since the conflict remains ongoing at the time of writing, this work can serve as an archival snapshot of model reasoning during an unfolding geopolitical crisis, enabling future studies without the hindsight bias of retrospective analysis.}

%% file: tex/introduction.tex
\section{Introduction}
\label{sec:introduction}

\textit{Would it have been possible to reasonably anticipate the outbreak of the Second World War (WWII) from within the 1930s, before its full escalation became historically obvious?} 
Questions of this kind are often discussed in hindsight, where the chain of events leading to a major conflict can appear almost inevitable. Yet this perception is strongly shaped by \textit{hindsight bias} \citep{fischhoff1975hindsight}. In reality, anticipating this kind of geopolitical events before they occur is extraordinarily difficult, even for experienced analysts and forecasters \citep{tetlock2016superforecasting, tetlock2017expert}. The challenge lies not merely in the scarcity of information, but in interpreting incomplete, ambiguous, and often contradictory signals in real time, without knowing which factors will ultimately prove decisive.

This historical thought experiment highlights a broader question about artificial intelligence (AI) reasoning in complex environments. Real-world geopolitical crises involve intertwined dynamics across military strategy, economic incentives, diplomacy, domestic politics, and human perception \citep{Betts1978AnalysisWA, Jervis1976PerceptionAM}. 
Effective reasoning in these environments demands the ability to navigate uncertainty, shifting incentives, and partial observability, which are the classic conditions often described as the \textit{fog of war}.

Understanding whether current state-of-the-art (SOTA) Large Language Models (LLMs) exhibit such capabilities is challenging. Retrospective evaluation of historical events is fundamentally confounded by training data leakage~\citep{carlini2021extractingtrainingdatalarge, Aiyappa_2023, Kang_2024}. 
Major geopolitical events are extensively documented in the vast corpora used to pretrain modern models \citep{brown2020languagemodelsfewshotlearners,Bender_parrot}, meaning that models may implicitly encode knowledge of outcomes even when prompted to reason from earlier points in time. 
As a result, retrospective prediction tasks can blur the distinction between genuine reasoning and latent memorization. 
Recent work has therefore raised growing concerns that many evaluation benchmarks may inadvertently measure pattern recognition or data leakage rather than true out-of-distribution reasoning ability \citep{magar2022datacontaminationmemorizationexploitation, sainz2023nlpevaluationtroubleneed, li2023estimatingcontaminationperplexityquantifying, yang2023rethinking}.

\begin{figure}[t]
    \centering
    \includegraphics[width=\textwidth]{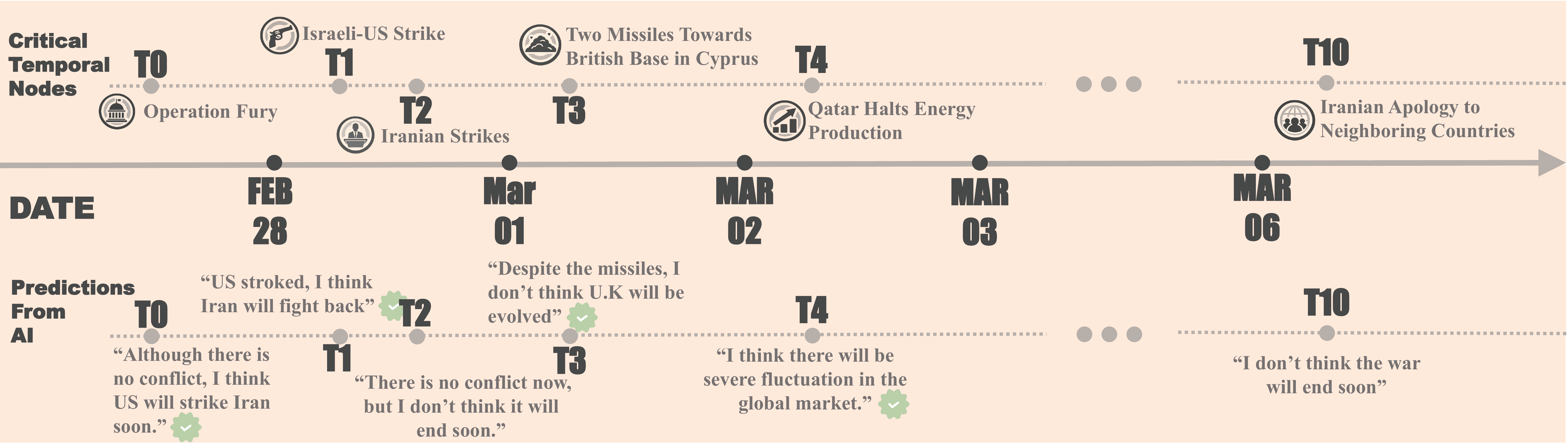}
        \caption{Critical temporal nodes during the early stages of the 2026 Middle East conflict (top) and representative model analyses at selected nodes (bottom). Models exhibit recognizable patterns of strategic reasoning under the fog of war, and their analyses evolve as new information becomes available across the timeline.
        }
    \label{fig:main}
\end{figure}

To study how LLMs reason under conditions that more closely resemble real-world uncertainty, we turn to a crisis that unfolded entirely after the training cutoff of current frontier models. 
\textit{The sudden escalation of the Middle East conflict in late February and early March 2026} \citep{acled2026middleeast} provides a rare opportunity to observe how language models interpret an unfolding geopolitical situation without access to the eventual outcomes. 
Because the early stages of this crisis occurred outside the training distribution of existing models, it offers \textit{a natural setting in which models must rely on their reasoning processes rather than learned knowledge} as shown in Figure \ref{fig:main}.

In this work, we use the early stages of the 2026 Middle East conflict as a temporally grounded case study for analyzing \textbf{how LLMs interpret and reason about an unfolding geopolitical crisis under strict information constraints}.
We reconstruct a timeline consisting of $11$ critical temporal nodes during the early stages of the conflict and formulate $42$ event-specific questions and $5$ general exploratory questions probing distinct aspects of geopolitical reasoning. 
The construction of these temporal nodes and questions was human-informed: in addition to reviewing contemporaneous reporting and public records, we incorporated input from individuals with lived experience of the war to improve the ecological validity of the timeline and the relevance of the reasoning probes. 
These questions span themes including \textit{Initial Outbreak}, \textit{Threshold Crossings}, \textit{Economic Shockwaves}, and \textit{Political Signaling}. 
At each temporal node, models are provided only with contextual information that would have been publicly available up to that moment and are asked to analyze potential developments and strategic implications.

\textit{Rather than} treating this setting as a forecasting benchmark, \textbf{our objective is to examine how language models behave when confronted with a complex, evolving real-world scenario}. In an ongoing conflict, many of the most important questions are not cleanly exhausted events with timeless binary labels: some concern degrees of escalation, some remain contingent on future developments, and some may be \textit{not yet} rather than definitively \textit{no}. By placing models within a temporally unfolding information environment, we are able to observe how they interpret uncertain signals, which factors they treat as strategically significant, and how their narratives evolve as additional information becomes available. This perspective allows us to study the qualitative reasoning patterns that emerge when LLMs attempt to analyze geopolitical dynamics under the conditions commonly described as the \textit{fog of war}.

\textbf{This work captures a snapshot of LLM reasoning during an unfolding geopolitical crisis, which remains ongoing at the time of writing. }
Unlike retrospective analyses of historical crises, the ultimate trajectory of the war, \textit{whether it stabilizes regionally, escalates further to a global war, or reaches a negotiated settlement}, has not yet been determined. 
Accordingly, even our ``verifiable'' questions should be understood as operational probes rather than immutable benchmark labels: they let us anchor parts of the analysis quantitatively, but they do not collapse the full ambiguity of the conflict into a final closed-world test set. To preserve this moment of uncertainty and reduce future hindsight distortion, this work archives the LLM responses produced at each temporal node as a record of contemporaneous LLM reasoning. 
As events continue to develop, these forecasts and narratives may serve as a reference point for future comparison and follow-up research.

\paragraph{Key Findings.}
\begin{itemize}
    \item \textbf{LLMs often show strong strategic reasoning under uncertainty.} Across multiple temporal nodes, model responses move beyond political rhetoric and focus instead on factors such as military sunk costs, deterrence pressures, and material constraints; in several early nodes, some models also anticipate escalation before kinetic conflict begins.

    \item \textbf{Their strengths are domain-specific rather than uniform.} Models are most reliable when reasoning about structural economic dynamics and material constraints, but less consistent in highly ambiguous political settings involving signaling, leadership instability, and multi-actor strategic interaction.

    \item \textbf{Their narratives evolve with the conflict unfolding.} As the conflict continues to unfold and new information becomes available, models move away from early expectations of rapid containment and increasingly converge on longer, more systemic accounts of the conflict.
\end{itemize}

\paragraph{Main Contributions.}
\begin{itemize}
    \item \textbf{A temporally grounded case study of LLM reasoning under the fog of war.}
    To our knowledge, this is among the first works to examine how LLMs \textit{analyze an unfolding war scenario} under strict temporal information constraints, where \textit{the outcome remains unknown and models must reason under real-time uncertainty}.

    \item \textbf{A structured framework for analyzing model reasoning in this unfolding scenario.}
    We construct a timeline of $11$ critical temporal nodes and design $42$ reasoning probes spanning military escalation, economic shockwaves, and political signaling, \textit{enabling longitudinal observation of how model analyses evolve as new information becomes available}.

    \item \textbf{An archived snapshot of LLM reasoning without final outcome.}
    We preserve the model responses generated at each temporal node as a record of reasoning under real-time uncertainty, \textit{providing a reference point for future research and retrospective comparison as the conflict continues to develop}.
\end{itemize}

\paragraph{Broader Impacts.} \leavevmode\\
This work has both constructive potential and associated risks. On the constructive side, studying how AI reasons under real-world geopolitical uncertainty may help researchers better understand both the capabilities and limitations of these systems in complex, high-stakes environments. The archived record of model reasoning generated during unfolding events may also provide a useful resource for future research on temporal reasoning and narrative evolution in AI systems.
At the same time, research on AI reasoning in geopolitical contexts remains preliminary, and the outputs of such systems should be interpreted with caution. This work is intended for analytical and research purposes rather than operational or military use. By examining how AI systems interpret incomplete signals and potential escalation dynamics, we aim to support research on forecasting, conflict prevention, and analytical transparency.
More broadly, we hope that improving our understanding of AI reasoning about geopolitical events can contribute to research aimed at reducing misinterpretation, mitigating escalation risks, and ultimately supporting efforts toward de-escalation and peace.

%% file: tex/background.tex
\section{Background and Related Work}
\label{sec:background}

\subsection{LLMs in Geopolitical Forecasting}

LLM forecasting has attracted growing interest as a setting for studying complex reasoning under real-world uncertainty~\citep{halawi2024approaching, karger2024forecastbench}. MIRAI~\citep{ye2024mirai} evaluates LLM agents over structured event databases for short- to long-horizon geopolitical prediction, while ForecastBench~\citep{karger2024forecastbench} shows that LLMs still substantially underperform expert human forecasters on unresolved future questions. EvolveCast~\citep{yuan2025assessing} examines how models update forecasts in response to new evidence and finds that revisions are typically overly conservative and inconsistent.
In political domains specifically, UNBench~\citep{liang2025benchmarking} targets UN Security Council vote prediction, and ThinkTank-ME~\citep{li2026thinktank} introduces a Middle East-focused event forecasting benchmark arguing for multi-expert collaboration.

Two important methodological concerns thread through this literature. \citet{paleka2025pitfalls} identify temporal leakage as a persistent confound in geopolitical forecasting evaluations, and \citet{li2026simulated} show that simply prompting models to ``pretend not to know'' pre-cutoff outcomes does not reliably simulate true ignorance.
These critiques directly motivate our design: by anchoring the study to a conflict that postdates all current model training, we ensure that neither parametric recall nor simulated ignorance can substitute for genuine real-time reasoning. Where prior work asks whether models can predict outcomes, we pursue a harder question: \textit{can models reason coherently about a crisis as it unfolds, armed only with the partial, noisy information available at each moment?}

\subsection{LLMs in Multi-Actor Social and Strategic Reasoning}

A parallel line of work examines whether LLMs can track the beliefs, intentions, and incentives of multiple agents simultaneously.
Theory-of-mind (ToM) benchmarks establish that while LLMs perform well on simplified belief-attribution tasks, they degrade on settings requiring multi-step mental-state tracking, hidden information, or higher-order recursive reasoning~\citep{gandhi2023understanding, kim2023fantom, wu2023hi}.
More recent work shifts from static ToM tests toward interactive and strategic settings: SOTOPIA~\citep{zhou2023sotopia} uses open-ended role-play to evaluate social goal coordination, SPIN-Bench~\citep{yao2025spin} probes strategic reasoning under incomplete information and multi-agent negotiation, and Mirofish~\citep{mirofish2025} constructs high-fidelity agent societies from real-world seed data to simulate collective social evolution and forecast future outcomes.

Geopolitical crises represent an extreme case of this multi-actor reasoning challenge: they involve numerous state and non-state actors with conflicting incentives, cascading second-order effects, and rapidly shifting informational landscapes.
Existing benchmarks either simplify the actor space (e.g., two-player games) or treat geopolitical scenarios as static prediction tasks.
Our test explicitly probes \textbf{multi-actor political signaling}, \textbf{threshold-crossing dynamics}, and \textbf{economic spillover reasoning}: capabilities that demand precisely the kind of extended, context-sensitive, multi-agent world modeling that current evaluations leave largely untested.

\subsection{LLM Reasoning Evaluation}

Standard reasoning benchmarks (e.g., MMLU~\citep{hendrycks2020measuring}, GSM8K~\citep{cobbe2021training}, BBH~\citep{suzgun2023challenging}, GPQA~\citep{rein2024gpqa}) treat reasoning as the solution of decontextualized problems with fixed inputs and predefined answer spaces.
More recent work has enriched evaluation by incorporating heterogeneous evidence, including charts, tables, and multimodal documents~\citep{yue2024mmmu, ma2024mmlongbench}, or by testing retrieval over novel, multi-source inputs~\citep{li2025towards, chen2025visr}.
A more recent line of work moves further toward real-world conditions: CaughtCheating~\citep{li2025caughtcheating} requires models to infer socially situated implications from weak visual cues, while forecasting-oriented benchmarks evaluate reasoning over unresolved future events~\citep{halawi2024approaching, karger2024forecastbench, yuan2025forecast}. Despite this progress, even the most grounded of this work still presents reasoning instances as static snapshots; the model is given a fixed context and asked to produce an answer. None of them track how reasoning \textit{evolves} as new information arrives incrementally over time.
Our study introduces a distinctive temporal constraint: models receive only information available at each of 11 sequential decision points and are repeatedly asked to update their analysis as the crisis unfolds, enabling us to examine not only answer accuracy but also belief revision and narrative coherence under the fog of war.

\subsection{Data Leakage in LLM Evaluation}

Data leakage in LLM evaluation goes far beyond simple train-test overlap. Given the scale and heterogeneity of modern pretraining corpora, leakage has become a systematic, multi-stage threat to reliable evaluation~\citep{deng2023benchmark, xu2024benchmark, cheng2025survey}. Critically, leakage is not limited to verbatim reproduction: paraphrased or translated benchmark items can evade standard decontamination while still inflating scores~\citep{yang2023rethinking}, and leakage can even cross language barriers and remain invisible to surface-overlap detectors~\citep{yao2024data}.
Empirical audits have found leakage levels ranging from 1\% to 45\% across popular QA benchmarks, with contamination growing over time~\citep{li2024open}. These findings collectively undermine the common assumption that benchmark scores constitute clear evidence of reasoning ability.

One mitigation strategy is \textit{dynamic benchmark design}: LatestEval \citep{li2024latesteval} sources questions from recent corpora, LiveBench \citep{white2024livebench} refreshes tasks on a rolling schedule, and LiveCodeBench \citep{jain2024livecodebench} continuously collects newly released programming problems. However, \citet{sun2025emperor} demonstrates that most existing mitigation strategies still fail to jointly preserve evaluation fidelity and contamination resistance. Our work takes a stricter approach. Rather than refreshing test items, we study model reasoning on a geopolitical crisis that unfolded \textit{entirely after} the training cutoff of all evaluated models, and we additionally restrict each query to information available only up to a specific temporal node. This substantially reduces not just verbatim leakage but also retrospective knowledge from paraphrased or cross-lingual contamination, making it among the most leakage-resistant evaluation settings currently feasible.

%% file: tex/method.tex
\section{Study Design}
\label{sec:method}

\subsection{Critical Temporal Nodes Construction}

To study how language models reason about unfolding real-world events, we aim to construct and select a timeline of critical temporal nodes representing key turning points during the early stages of the conflict.
Each temporal node corresponds to a moment at which new information substantially alters the strategic landscape, such as the initiation of military operations, retaliatory strikes, escalation involving additional actors, or major political and economic developments.

Formally, we define each temporal node $T_i$ as a snapshot of the information environment available at a specific time. For every node, we compile a contextual information package consisting of publicly reported news available up to that moment, which we use as the input context $C_{T_i}$ for the language model.
Crucially, we do not include any information published after $T_i$ in the context, \textbf{ensuring that model responses do not rely on knowledge of future outcomes.} Because the conflict unfolded after the training cutoff of all evaluated models, the risk of training-data leakage is substantially reduced, making this a setting well-suited for studying reasoning under genuine uncertainty.

\input{table/timeline_nodes_table.tex}

To ensure that the selected nodes reflect not only formal geopolitical milestones but also the moments perceived as most consequential by people directly following the events, \textbf{we conduct informal interviews with five individuals located in the Middle East during the early stages of the conflict.} Participants recall the moments they remember most vividly since the beginning of the war, as well as the events that most change their perception of the conflict's trajectory.
We combine these perspectives with a systematic review of publicly reported developments across international news sources in our selection process.

The final resulting timeline contains $11$ critical temporal nodes as shown in Table \ref{tab:temporal_nodes}.
Together, these nodes capture multiple themes of geopolitical dynamics, including \textbf{Initial Outbreak}, \textbf{Threshold Crossings}, \textbf{Economic Shockwaves}, and \textbf{Political Signaling}. By structuring the analysis around these nodes, we approximate a sequence of real-time reasoning scenarios in which both humans and language models interpret incomplete information and anticipate potential developments under uncertainty.

\subsection{Question Design}

\subsubsection{Node-Specific Verifiable Questions}

For each temporal node, we design a set of general event-based questions to probe how language models reason about real-world developments. \textbf{Across $11$ temporal nodes, the study includes $42$ verifiable questions}, with each node associated with three to five questions focusing on broad developments rather than specific details.

The process of designing these questions was informed by \textbf{two complementary sources}.
First, during the aforementioned interviews, participants were asked about the types of developments or turning points they paid attention to and what possibilities or general trends they wondered about as the events progressed.
Second, we examined news reports and policy analysis from the early stages of the crisis to summarize the kinds of overall uncertainties and broader topics discussed by commentators and analysts.

Drawing on these perspectives and incorporating a thematic framework, we constructed a series of questions that focus on event trends, \textbf{aiming to understand how language models reason about and analyze the progression of major events as they unfold}. These questions primarily concern developments whose outcomes can later be compared against real-world events to obtain the ground truth. The detailed questions are listed within each theme.

At the same time, we do not view these questions as forming a conventional benchmark in which every item admits a perfectly final, context-independent binary label. In an ongoing conflict, some developments are best understood as matters of degree, and even apparently binary outcomes may remain temporally open: an event that has not occurred yet at the time of analysis could still occur later. We therefore use these questions as structured analytical anchors that make portions of the evolving situation more comparable across models, while keeping the primary emphasis on the reasoning trajectories reflected in the responses.

\subsubsection{General Exploratory Questions}

In addition to these verifiable questions, we also include \textbf{$5$ general exploratory questions} that are asked at every temporal node $T_i$. These questions concern broader geopolitical trajectories that are inherently difficult or impossible to verify directly within the scope of the study. Specifically, these general questions are listed in Table \ref{tab:general_questions}.

\input{table/general_questions_table.tex}

Unlike the node-specific questions, responses to these exploratory questions are \textit{not evaluated for correctness}. Instead, they serve a different purpose: \textbf{to record how language models interpret and narrate the evolving situation as the conflict unfolds.} By asking the same set of broad questions at each temporal node $T_i$, we obtain a longitudinal record of how model analyses change in response to new information.

We include these responses as part of the released dataset, providing a structured archive of model interpretations during an unfolding geopolitical crisis. We hope that this record may support future research on temporal reasoning, narrative evolution in language models, and the study of how AI systems interpret complex real-world events over time.

\subsection{Interaction Protocol}
\label{sec:interaction_protocol}

We situate language models within a temporally grounded reasoning protocol defined by the temporal nodes $T_i$, the contextual information packages $C_{T_i}$, and the corresponding questions. Each question is posed independently.

For every question associated with a temporal node $T_i$, the model receives the complete contextual information package $C_{T_i}$ as input. The context consists of publicly reported news articles and their sources that were available up to time $T_i$. Importantly, the context only contains \textbf{information preceding the temporal node}, but never includes any information published after $T_i$, ensuring that the model does not have access to future developments.

Each question is asked independently using the following prompt structure:

\begin{quote}
\textit{\{context\}}

Based on the above publicly available information, try to analyze the current situation and potential future direction, then respond to this question: \textit{\{question\}}

At the end of your response, also provide the probability.
\end{quote}

This prompting format encourages models to first reason about the geopolitical situation implied by $C_{T_i}$ and then provide an explicit probability estimate for the event described in the question. We record the full model response, including both the reasoning process and the final probability assessment.

In practice, the probability estimates were not always expressed in a perfectly standardized numerical format. To ensure accurate quantitative analysis, \textit{two researchers independently read all the model responses and manually extracted the final probability judgment for each node-specific verifiable question.}
Disagreements were resolved through discussion.
We adopted this manual extraction procedure because it was more reliable than automatic parsing for preserving the probability actually implied by the model's final response.

For the node-specific verifiable questions, we operationalize ground truth with respect to a fixed observation cutoff: an event is labeled as having occurred if it had occurred by the time of writing, and not occurred otherwise. In the current study, we give each question at least a one-week window for resolution. This is a pragmatic operationalization rather than a claim that one week is uniquely correct: in an ongoing geopolitical conflict, some event labels may shift under a shorter or longer observation window. The value of this choice is therefore transparency and reproducibility for limited quantitative comparison, while acknowledging that in an ongoing conflict \textit{not yet observed} should not be over-interpreted as an absolute negation. 

As a consequence, our primary goal is \textit{not to treat this task as a conventional benchmark}, but rather to analyze how language models interpret incomplete real-world information and construct narratives about unfolding geopolitical events. Therefore, \textbf{our analysis focuses primarily on the qualitative reasoning patterns present in model responses, including how models interpret strategic incentives, escalation dynamics, and cross-domain implications}, with the quantitative labels serving as auxiliary signals rather than definitive final judgments.

\subsection{Context Corpus}

We construct a temporally structured context corpus spanning $11$ nodes that track the escalation of the conflict from Feb.~01 to Mar.~07, 2026. Articles are sourced from $12$ international outlets covering diverse regional and editorial perspectives, including \textit{Middle East Eye}, \textit{Al Jazeera}, \textit{The National (UAE)}, \textit{Al-Monitor} for regional coverage, and \textit{Reuters}, \textit{Bloomberg}, \textit{The Guardian}, \textit{Financial Times}, \textit{AP News}, \textit{BBC}, and \textit{Fox News} for Western perspectives. We collected this news via Google News RSS feeds and direct outlet links starting from Feb.~01, 2026, which marks the earliest available coverage of the conflict's onset. For articles where the RSS feed returns only a headline, we resolve the redirect URL and extract the full article body using \texttt{trafilatura}~\cite{barbaresi-2021-trafilatura}. Duplicates are removed via normalized title matching. For each temporal node $T_i$, the context window is built by filtering all articles published \emph{strictly before} the corresponding event timestamp, then sorting them in reverse chronological order (most recent first). Each article is formatted as \texttt{[YYYY-MM-DD HH:MM] Title (Source)} followed by up to 2,000 characters of body text if substantive content is available beyond the headline; otherwise, only the headline line is included. Entries are concatenated sequentially until the total exceeds 480,000 characters ($\approx$120K tokens), at which point the remaining older articles are dropped and replaced with a single \texttt{[...N older articles omitted]} marker. The resulting context block is then prepended to the prompt template described in Section~\ref{sec:interaction_protocol}.

\begin{figure}[t]
    \centering
    \includegraphics[width=\textwidth]{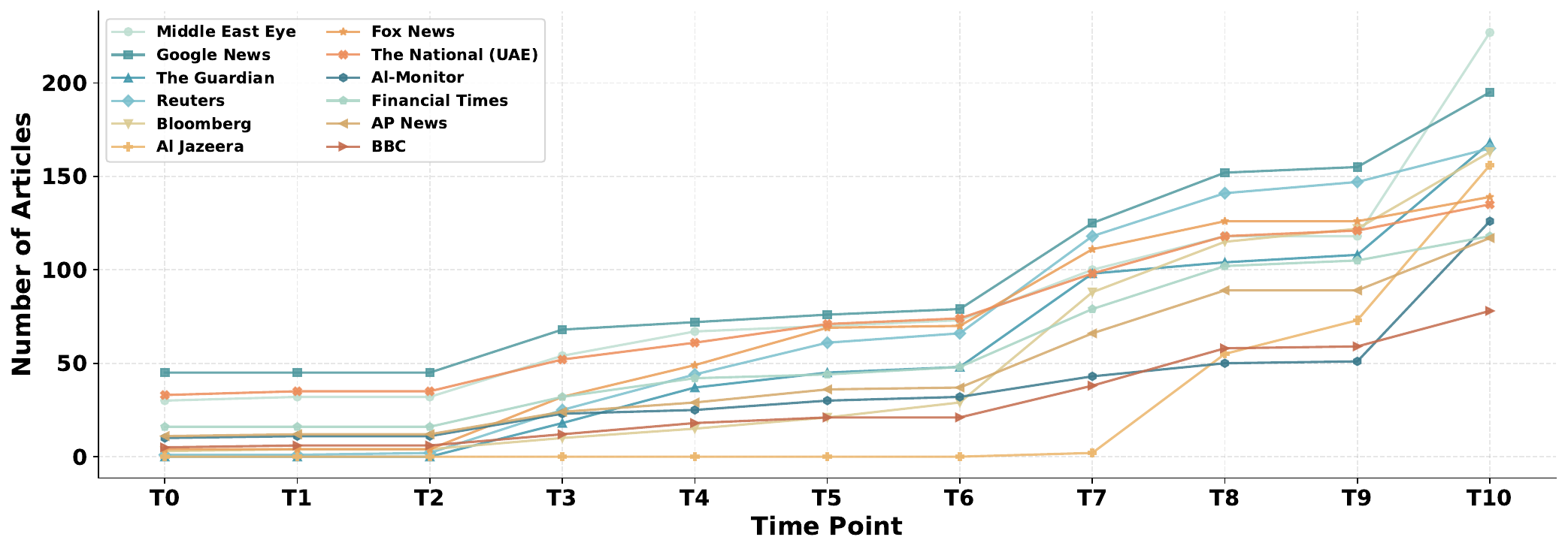}
        \caption{\textbf{Distribution of news articles across sources over time.} The figure plots the number of news articles published by major media outlets at each temporal node.}
    \label{fig:source_distribution}
\end{figure}

Critically, we apply \textbf{no further filtering, summarization, or editorial curation}: multiple outlets reporting on the same event are all retained, and syndicated or updated articles are treated independently. This is intentional; it simulates the raw, unstructured information environment a real-world forecasting agent would encounter, where redundancy, noise, and conflicting reports are natural features rather than artifacts to be removed. The cumulative article pool grows from 158 articles at $T_0$ to 1,787 at $T_{10}$ (totaling $\approx$889K words), with per-source growth shown in Figure~\ref{fig:source_distribution}.

%% file: table/timeline_nodes_table.tex
\begin{table}[t]
\centering
\caption{The selected $11$ critical temporal nodes with their corresponding themes.}
\label{tab:temporal_nodes}
\resizebox{\textwidth}{!}{%
\begin{tabular}{cllcl}
\toprule
\textbf{Node} & \textbf{Date} & \textbf{Event} & \textbf{Theme} & \textbf{Theme Disruption} \\
\midrule
T0 & Feb 27 & Operation Epic Fury & I & Initial Outbreak \\
T1 & Feb 28 & Israeli-US Strikes & I & Initial Outbreak \\
T2 & Feb 28 & Iranian Strikes & I & Initial Outbreak \\
T3 & Mar 1 & Two Missiles towards British Bases on Cyprus & II & Threshold Crossings \\
T4 & Mar 1 & Oil Refiner and Oil Tanker Was Attacked & III & Economic Shockwaves \\
T5 & Mar 2 & Qatar Halts Energy Production & III & Economic Shockwaves \\
T6 & Mar 2 & Natanz Nuclear Facility Damaged & II & Threshold Crossings \\
T7 & Mar 3 & U.S. Begins Evacuation of Citizens from the Middle East & II & Threshold Crossings \\
T8 & Mar 3 & Nine Countries Involved and Israeli Ground Invasion & II & Threshold Crossings \\
T9 & Mar 3 & Mojtaba Khamenei Becomes Supreme Leader & IV & Political Signaling \\
T10 & Mar 6 & Iranian Apology to Neighboring Countries & IV & Political Signaling \\
\bottomrule
\end{tabular}%
}
\end{table}

%% file: table/general_questions_table.tex
\begin{table}[t]
\centering
\caption{General Exploratory Questions.}
\label{tab:general_questions}
\begin{tabular}{@{}r@{\hspace{1em}}p{\dimexpr\textwidth-4em\relax}@{}}
\toprule
\textbf{\#} & \textbf{General Exploratory Questions} \\
\midrule
1 & What are the potential future actions by the United States and Israel? \\
2 & What are the potential future actions by Iran? \\
3 & What are the potential involvement or reactions from other major countries? \\
4 & Will the conflict escalate into a global war? \\
5 & What is the most probable pathway to de-escalation or resolution of the Iran-US conflict, and what is a realistic timeline? \\
\bottomrule
\end{tabular}
\end{table}

%% file: tex/main_part.tex
\section{Reasoning Analysis under the Fog of War}
\label{sec:thematic_analysis}

We begin by examining the qualitative reasoning trajectories that emerged when models interpreted the unfolding conflict under the fog of war. 
To capture these nuances, our analysis is based on manual reading of the full model responses generated for each question at every temporal node.
Rather than relying on automated keyword extraction, we inductively categorized the inferential trajectories and narrative shifts present in the generated text.

Throughout this section, we provide representative response snippets to illustrate the behavior of the models. All model responses are presented in blockquote format and represent the original text generated by the models. 

\subsection{General Model Behaviors and Stylistic Differences}

Before examining individual temporal nodes in detail, several general patterns emerge from comparing the reasoning styles of the evaluated models. For a better understanding of the model's reasoning trajectories, we provide some complete examples in Appendix~\ref{appendix:complete_examples}.

First, most models display a strong tendency toward structural strategic reasoning rather than simply repeating political rhetoric. Across multiple questions, models commonly grounded their analysis in factors such as military deployments, deterrence dynamics, and institutional incentives. 

Second, the models differ in how structured and formalized their reasoning narratives tend to be. Some models, particularly Claude and GPT-5.4, often produce highly structured analyses that explicitly separate drivers, constraints, and scenario pathways. Their responses frequently organize reasoning into sections such as escalation indicators, limiting factors, and alternative outcomes. By contrast, models such as Gemini and Qwen more often construct broader narrative explanations, integrating political signals, regional dynamics, and economic implications into a single analytical storyline.

A third difference concerns the level of analytical granularity. Some models tend to enumerate detailed operational scenarios and probability breakdowns, outlining multiple escalation pathways or target sets. Others provide shorter assessments focused on the most probable strategic trajectory without extensive scenario decomposition. These stylistic differences affect the length and structure of the responses, but they generally do not lead to radically different overall judgments.

\input{tex/main_theme_1.tex}
\input{tex/main_theme_2.tex}
\input{tex/main_theme_3.tex}
\input{tex/main_theme_4.tex}

%% file: tex/main_theme_1.tex
\subsection{Case Study on Theme I: Initial Outbreak \& Military Spillover}
\input{table/theme1_questions_table.tex}

\subsubsection{T0: Feb 27, Operation Epic Fury}

The first temporal node (T0) revealed evidence of early-stage geopolitical reasoning even before any military conflict had occurred. \textbf{Across models, the dominant framing was that the situation had moved beyond routine coercive signaling and toward open military conflict.}
In this section, we focus our analysis of T0 on two inferential patterns: \textit{how models reasoned about strategic sunk costs, and how they diverged when processing diplomatic signals.}

\paragraph{Strategic Intuition and the ``Credibility Trap''.} 
When evaluating the likelihood of direct military strikes, the models showed evidence of reasoning about structural incentives rather than merely reflecting surface-level rhetoric. Models such as \texttt{claude-sonnet-4.6} and \texttt{gemini-3.1-flash} frequently framed the situation as one in which the \textbf{sheer scale of the military deployment had created a point of no return}, emphasizing that logistical momentum in high-stakes geopolitics can override last-minute diplomatic hesitation.

\begin{modelquote}
\textbf{claude-sonnet-4.6}: ``The scale of the US deployment is \underline{extraordinary by any historical measure}... This is \underline{not merely signaling}. Deployments of this magnitude create their own political logic... The buildup has become \underline{a trap of Trump's own making}.''
\end{modelquote}

\begin{modelquote}
\textbf{gemini-3.1-flash}: ``The Trump administration has established a pattern of using military force... This has created a situation where the U.S. military buildup... is widely viewed by regional observers as a genuine preparation for \underline{a 'highly kinetic' campaign rather than mere posturing}.''
\end{modelquote}

This pattern of responses suggests a shared inferential baseline: \textbf{across models, the prevailing reasoning was that withdrawing such a massive force without extracting major concessions would result in a catastrophic loss of credibility.}

\paragraph{The Divergence: Susceptibility to Diplomatic Smokescreens.} 
However, this inferential consistency broke down when models were subsequently asked whether the U.S. might prioritize economic sanctions and diplomatic pressure over military action. \textbf{Despite all models acknowledging the imminence of strikes in the previous prompt, they weighted the conflicting signals of the ongoing Geneva talks very differently. }

Models such as \texttt{gpt-5.4} and \texttt{gemini-3.1-flash} responded quite differently. Because the contextual data included reports of ``significant progress'' from Omani mediators, these models heavily weighted the diplomatic signals, assigning higher probabilities to the U.S. \textbf{prioritizing sanctions rather than military action}. They tended to interpret the military buildup primarily as a leverage tactic rather than a commitment to strike.

\begin{modelquote}
\textbf{gpt-5.4}: ``Multiple reports say US-Iran talks in Geneva produced at least some 'progress'... That strongly suggests \underline{Washington has not yet decided diplomacy is exhausted}... Trump appears to \underline{want leverage, not necessarily war for its own sake}.''
\end{modelquote}

\begin{modelquote}
\textbf{gemini-3.1-flash}: ``\underline{The most likely path is a 'coercive diplomatic' outcome}. The U.S. is likely to maintain its military posture \underline{as a permanent threat to ensure compliance}, while pushing for a 'narrow' deal...''
\end{modelquote}

In contrast, \texttt{claude-sonnet-4.6} assigned a substantially lower probability to the diplomatic path, maintaining a more consistent inferential position across the two questions and \textbf{explicitly reasoning that sanctions could not remain the primary strategy at this stage.}

\begin{modelquote}
\textbf{claude-sonnet-4.6}: ``The US \underline{cannot realistically prioritize pure economic sanctions or diplomacy} as its primary strategy at this stage — the military buildup has created its own momentum and credibility requirements.''
\end{modelquote}

Taken together, T0 shows that models varied in how they weighed competing signals in the pre-conflict information environment. \textbf{Most models tended to frame the situation as having moved beyond routine coercive signaling and toward open military escalation.} Their main divergence lies not in whether they registered the risk of war, but in how much weight their outputs assigned to the residual diplomatic signals still present in the information environment. Some models interpreted the ongoing negotiations as evidence that diplomacy remained viable, while others emphasized the structural momentum created by the massive military buildup.

\subsubsection{T1: Feb 28, Israeli-US Strikes}

T1 marked the moment when the crisis transitioned from anticipatory tension into actual kinetic action.
At T1, the evaluation shifted from assessing hypothetical risks to analyzing active escalation ladders. \textbf{Despite the inherent chaos of the ``fog of war'', the models exhibited relatively stable reasoning patterns across this temporal node.} In this section, we highlight two inferential tendencies that emerged from the responses: \textit{an apparent resistance to historical bias, and a tendency to disentangle extreme political rhetoric from assessments of military doctrine.}

\paragraph{Overcoming Historical Bias.} 
A common pitfall in geopolitical forecasting is over-relying on recent historical precedents. The provided contextual information noted a previous ``12-Day War'' in June 2025, where strikes remained relatively contained and Iran's retaliation was highly restrained. However, \textbf{when evaluating whether the current strikes would escalate into a broader campaign, several models showed evidence of reasoning beyond this historical precedent. Their responses emphasized that the underlying strategic calculus had changed in ways that the June 2025 template could not capture.}

\begin{modelquote}
\textbf{claude-sonnet-4.6}: ``The 'June 2025 Template' Problem... the June 2025 'Midnight Hammer' operation was essentially a one-off strike \underline{where Iran responded weakly}... This time, the situation is fundamentally different... Iran's leadership has been \underline{liberated from its fear of total war} by having already survived one US attack.''
\end{modelquote}

\begin{modelquote}
\textbf{gemini-3.1-flash}: ``Iran's Shift in Calculus... Having already survived a significant degradation of their nuclear and military infrastructure, the regime seems to have \underline{moved past the 'paralyzing caution'} that previously deterred them.''
\end{modelquote}

\textbf{Rather than extrapolating mechanically from the earlier episode, the models tended to emphasize a psychological and strategic phase shift}, reasoning that the threshold for a broader, sustained military campaign had been crossed.

\paragraph{Disentangling Rhetoric from Rational Doctrine.} 
A further inferential pattern emerged when models were asked whether Iran's retaliation would likely involve indiscriminate bombing of Israeli-U.S. civilian territory rather than targeting military bases. At this node, the information environment was saturated with extreme rhetoric from Iranian officials threatening a ``regional war'' and ``severe'' consequences without limits. Yet, \textbf{the models generally discounted the inflammatory posturing.} They tended to reason that retaliation would remain calibrated toward military assets.

\begin{modelquote}
\textbf{gpt-5.4}: ``Iran's doctrine historically \underline{favors calibrated, politically framed retaliation}... Iran generally tries to present retaliation as: \underline{state-to-state, military-for-military, proportionate} or framed as defensive...''
\end{modelquote}

\begin{modelquote}
\textbf{claude-sonnet-4.6}: ``Escalation Management Incentives... Iran has consistently sought to \underline{avoid triggering} \underline{overwhelming US retaliation that could threaten regime survival}. Indiscriminate bombing... would virtually guarantee catastrophic US escalation, potentially including regime decapitation strikes.''
\end{modelquote}

Taken together, T1 illustrates that, when equipped with sufficient contextual information, the models showed evidence of reasoning about structural incentives rather than merely reflecting the surface-level content of the information environment. \textbf{Across responses, a recurring pattern was the tendency to discount emotional political rhetoric and contextualize historical analogies when assessing the mechanics of state survival and deterrence.}

\subsubsection{T2: Feb 28, Iranian Strikes}

As the conflict erupted into active warfare, T2 introduced profound geopolitical shocks, including reports of a pre-emptive U.S.-Israeli strike and the sudden death of Iran's Supreme Leader, triggering an immediate succession crisis. \textbf{Despite the influx of chaotic and potentially destabilizing variables, the models tended to navigate this escalation by linking Iran's domestic political fragility to its external military posture.} 

\paragraph{Leadership Transition and the Escalation Calculus.} 
When analyzing whether Iran's retaliation would remain merely ``symbolic'' with no sustained operations, the models showed evidence of reasoning about authoritarian regime survival logic. Rather than treating a sudden leadership vacuum as inherently paralyzing, \textbf{the models frequently framed the immediate succession of a new leader as a potential catalyst for a more aggressive response, reasoning that an untested leader faces strong incentives to avoid appearing weak.}

\begin{modelquote}
\textbf{claude-sonnet-4.6}: ``The appointment of Mojtaba Khamenei under Israeli threat creates a uniquely dangerous dynamic. The new supreme leader must: \underline{1. Establish credibility with IRGC hardliners}, \underline{2. Avoid appearing to capitulate}... This creates unpredictability - the new leadership may \underline{authorize more aggressive action... precisely to establish authority}.''
\end{modelquote}

\begin{modelquote}
\textbf{gpt-5.4}: ``Iran's leadership is under pressure: internal unrest... leadership transition/succession issues... In that environment, \underline{the regime likely feels it must show it can still impose costs}. Even if it wants to avoid total war, it also cannot appear unable to respond.''
\end{modelquote}

Taken together, T2 illustrates that the models did not uniformly associate leadership decapitation with regime collapse or military passivity. \textbf{A recurring inferential pattern was that existential external threats, when compounded by a succession crisis, were framed as creating strong pressures for authoritarian regimes to escalate externally in order to secure internal legitimacy.}

%% file: table/theme1_questions_table.tex
\begin{table}[t]
\centering
\caption{The Node-Specific Questions for Theme I: Initial Outbreak \& Military Spillover (T0, T1, T2).}
\label{tab:theme1_questions}
\begin{tabular}{@{}rp{\dimexpr\textwidth-5em\relax}@{}}
\toprule
\textbf{Idx} & \textbf{Node-Specific Questions} \\
\midrule
\multicolumn{2}{@{}l}{\textbf{T0: Feb 27, Operation Epic Fury}} \\
\midrule
1 & Is the United States likely to conduct direct military strikes on Iranian territory? \\
2 & Could the United States prioritize economic sanctions or diplomatic pressure over military action? \\
3 & Does the announcement of Operation Epic Fury signal the beginning of a broader military conflict? \\
\midrule
\multicolumn{2}{@{}l}{\textbf{T1: Feb 28, Israeli-US Strikes}} \\
\midrule
1 & Are the Israeli-U.S. strikes likely to escalate into a broader military campaign? \\
2 & Is Iran likely to retaliate militarily in response to the strikes? \\
3 & If Iran retaliates, is it more likely to indiscriminately bomb Israeli-U.S. territory than to target U.S. military bases? \\
4 & Could Gulf states become direct targets of Iranian retaliation \\
5 & Is regional airspace in the Middle East likely to face widespread closures due to security risks? \\
\midrule
\multicolumn{2}{@{}l}{\textbf{T2: Feb 28, Iranian Strikes}} \\
\midrule
1 & Are the Iranian strikes likely to remain symbolic retaliation without sustained military operations? \\
2 & Could Iran attempt to disrupt shipping through the Strait of Hormuz? \\
3 & Are other countries in the region likely to become directly involved in the conflict? \\
4 & Is large-scale closure of Middle Eastern airspace likely following these strikes? \\
5 & Could the conflict trigger internal rebel army within Iran? \\
\bottomrule
\end{tabular}
\end{table}

%% file: tex/main_theme_2.tex
\subsection{Case Study on Theme II: Threshold Crossings \& Internationalization}
\input{table/theme2_questions_table.tex}

\subsubsection{T3: Mar 1, Two Missiles towards British Bases on Cyprus}

T3 examined how models assessed the risk of internationalization. \textbf{In this node, two patterns emerged: evidence of reasoning about institutional mechanics (e.g., NATO protocols), alongside divergence in how models weighted domestic political noise.}

\paragraph{Institutional Realism.} 
When asked if NATO would formally enter the conflict, expanding it into the Mediterranean theater, a naive reading of the context might suggest imminent alliance-wide involvement. However, \textbf{the models showed evidence of reasoning about institutional constraints, distinguishing individual member state actions from NATO's collective defense mechanisms.}

\begin{modelquote}
\textbf{claude-sonnet-4.6}: ``NATO operates on consensus, and several members (Turkey, Hungary) would likely block formal alliance involvement... \underline{Article 5 collective defense provisions are not triggered} - no NATO member has been attacked by Iran directly.''
\end{modelquote}

\begin{modelquote}
\textbf{gpt-5.4}: ``NATO's collective-defense mechanism normally requires an attack on a member... \underline{So far, this is a US-Israel-Iran war, not clearly a NATO war}... Even if many NATO members are active, they may avoid calling it a NATO mission to prevent alliance splits.''
\end{modelquote}

\paragraph{The Distraction of Domestic Political Noise.} 
However, predicting the actions of individual states proved more challenging, revealing a divergence in how models weight domestic political rhetoric. When evaluating whether the United Kingdom would directly join the offensive, the context included loud demands from opposition politicians (e.g., Nigel Farage) urging the UK to join the U.S. strikes. 

Models like \texttt{claude-sonnet-4.6} and \texttt{gpt-5.4} tended to prioritize the structural reality that the UK Navy had withdrawn ships from the Gulf and explicitly opted out of the initial strike wave. Conversely, \texttt{gemini-3.1-flash} \textbf{assigned a markedly higher probability of involvement, appearing to overweight the vocal political opposition relative to the military indicators.}

\begin{modelquote}
\textbf{claude-sonnet-4.6}: ``The UK explicitly chose \underline{not to participate in initial strikes despite advance notice}... Royal Navy at historic low with no warships in the Arabian Gulf... The UK is most likely to remain in a supporting role.''
\end{modelquote}

\begin{modelquote}
\textbf{gemini-3.1-flash}: ``The likelihood of the United Kingdom becoming directly involved... is \underline{moderate to high}... driven by Domestic Political Pressure: There is significant and vocal pressure from opposition parties... Reform UK leader Nigel Farage and other MPs have publicly urged the government to 'back the Americans'.''
\end{modelquote}

Taken together, T3 suggested that models tended to reason more consistently when evaluating the formal constraints of international institutions like NATO. However, when assessing individual state actors, \textbf{the inferential outcomes varied considerably depending on whether models weighted hard military realities or domestic political rhetoric more heavily.}

\subsubsection{T6: Mar 2, Natanz Nuclear Facility Damaged}

At T6, the conflict escalated significantly with strikes on underground nuclear sites and the reported decapitation of Iran's leadership. \textbf{In a highly volatile information environment, the models tended to differentiate between the prospect of actual nuclear weapon deployment and the prospect of nuclear-related coercive signaling, assigning substantially higher probability to the latter.}

\paragraph{Deterrence Signaling vs. Actual Deployment.} 
When asked if nuclear weapons could be used as part of the conflict, the models tended to assess that while an actual nuclear detonation was highly unlikely, indirect nuclear threats, such as accelerated enrichment, diplomatic blackmail, or radiological incidents, were plausible coercive tactics. 
This pattern of responses illustrated evidence of reasoning about \textbf{the fundamental logic of nuclear strategy}: that in regional conflicts, nuclear programs function primarily as instruments of coercive signaling rather than battlefield deployment.

\begin{modelquote}
\textbf{claude-sonnet-4.6}: ``The most realistic nuclear-related threat is \underline{nuclear rhetoric}... rather than actual nuclear weapon deployment... The new leadership... faces a survival calculation where \underline{nuclear use would guarantee total destruction}.''
\end{modelquote}

\begin{modelquote}
\textbf{gpt-5.4}: ``The most plausible nuclear dangers are \underline{indirect}... Iranian 'breakout' threat: Iran could announce withdrawal from remaining nuclear commitments... That would be \underline{a nuclear crisis} \underline{even before any bomb exists}.''
\end{modelquote}

\paragraph{Overestimating Institutional Escalation.} 
However, this inferential consistency broke down when models were asked whether Iran would formally withdraw from its nuclear non-proliferation commitments. Faced with the extreme context of leadership decapitation, models like \texttt{gemini-3.1-flash} \textbf{appeared to over-extrapolate the escalation, treating military conflict as likely to immediately dissolve all institutional hedging.}

\begin{modelquote}
\textbf{gemini-3.1-flash}: ``The U.S. and Israel have explicitly stated that the goal of the current military campaign is regime change... In this environment, \underline{adherence to international treaties like the } \underline{Non-Proliferation Treaty (NPT) is viewed by the regime as a liability}...''
\end{modelquote}

Conversely, \texttt{claude-sonnet-4.6} offered a contrasting assessment, reasoning that \textbf{even in a survival scenario, formal NPT withdrawal carries catastrophic diplomatic costs} that constrain the regime's options.

\begin{modelquote}
\textbf{claude-sonnet-4.6}: ``Iran is \underline{unlikely to make a formal, explicit withdrawal} from the NPT in the near term... \underline{Formal withdrawal would invite immediate escalation}...''
\end{modelquote}

Taken together, T6 suggested that while models showed evidence of reasoning about the limits of direct nuclear use, \textbf{many tended to underweight the degree to which state actors maintained complex institutional commitments (like NPT membership) even during existential crises.}

\subsubsection{T7: Mar 3, U.S. Suggests Evacuation of Citizens from the Middle East}

At T7, the conflict caused massive regional disruptions, and U.S. officials suggested civilian evacuations. When asked if the U.S. might deploy ground forces, the models tended to identify a tension between the official preference for an air-only campaign and the operational requirements implied by the stated war objectives.

\paragraph{The Limits of Airpower versus Actual Goals.} 
U.S. officials publicly stated a preference for airstrikes to avoid a protracted conflict, while simultaneously refusing to rule out ground forces. Rather than treating these political statements as definitive, the models tended to reason from the stated war objectives, regime change, and securing nuclear materials, and frequently emphasized that such objectives are difficult to achieve through airpower alone.

\begin{modelquote}
\textbf{claude-sonnet-4.6}: ``Trump has explicitly refused to rule out ground troops... Several stated US objectives are \underline{difficult to achieve from the air alone}: Verifying destruction of nuclear facilities, Securing nuclear materials, Ensuring regime change takes hold.''
\end{modelquote}

\begin{modelquote}
\textbf{gpt-5.4}: ``The stated aims have broadened in public discussion: not just degrading missiles or nuclear capacity, but also leadership decapitation... \underline{That kind of widening objective set can create conditions} \underline{where airpower alone is judged insufficient}.''
\end{modelquote}

\begin{modelquote}
\textbf{gemini-3.1-flash}: ``The primary argument for ground forces would be the need to \underline{secure nuclear sites} \underline{or neutralize mobile missile launchers that air power alone has failed to eliminate}.''
\end{modelquote}

Taken together, T7 illustrated that models tended to reason beyond public political rhetoric when assessing military requirements. \textbf{A recurring inferential pattern was that air campaigns were unlikely to be sufficient if the stated objectives included regime change or the physical seizure of nuclear sites,} suggesting that models associated these goal-sets with an elevated probability of eventual ground force deployment.

\subsubsection{T8: Mar 3, Nine Countries Involved and Israeli Ground Invasion}

At T8, the conflict had already drawn in nine countries, and Israel had launched ground operations into Lebanon. When asked if the conflict would expand further, all models inferred high probabilities of further expansion.
\textbf{A common reasoning pattern was that neighboring countries would prefer to remain outside the conflict, yet existing military agreements function as tripwires that could draw reluctant states into the battlefield.}

\begin{modelquote}
\textbf{claude-sonnet-4.6}: ``\underline{Pakistan explicitly signaled its defense pact with Saudi Arabia could draw it in}... If Iran deliberately targets Turkish territory... \underline{Article 5 considerations could draw NATO more formally } \underline{into the conflict}.''
\end{modelquote}

Beyond formal treaties, the models also showed evidence of reasoning about the downstream consequences of the U.S. and Israeli decapitation strategy. Rather than treating leadership removal as likely to end the conflict quickly, \texttt{gemini-3.1-flash} emphasized the risk of a power vacuum that could draw neighboring states in.

\begin{modelquote}
\textbf{gemini-3.1-flash}: ``The primary factor that could trigger further expansion is the collapse of the Iranian state. \underline{If the regime fragments, the resulting chaos could draw in neighboring countries ( Turkey or Iraq)} \underline{to secure their borders}... effectively turning the conflict into a multi-front regional civil war.''
\end{modelquote}

Taken together, T8 illustrated that models tended to frame further conflict expansion not as driven by states eager to join the fighting, but as a product of \textbf{the structural triggers of defense pacts and the spillover effects of potential regime collapse}, a pattern of reasoning that emphasized involuntary entrapment over voluntary escalation.

%% file: table/theme2_questions_table.tex
\begin{table}[t]
\centering
\caption{The Node-Specific Questions for Theme II: Threshold Crossings \& Internationalization (T3, T6, T7, T8).}
\label{tab:theme2_questions}
\begin{tabular}{@{}rp{\dimexpr\textwidth-5em\relax}@{}}
\toprule
\textbf{Idx} & \textbf{Node-Specific Questions} \\
\midrule
\multicolumn{2}{@{}l}{\textbf{T3: Mar 1, Two Missiles towards British Bases on Cyprus}} \\
\midrule
1 & Is the United Kingdom likely to become directly involved in the conflict? \\
2 & Is NATO likely to become involved, expanding the conflict into the Mediterranean theater? \\
3 & Could the conflict disrupt commercial shipping or maritime security in the Mediterranean? \\
\midrule
\multicolumn{2}{@{}l}{\textbf{T6: Mar 2, Natanz Nuclear Facility Damaged}} \\
\midrule
1 & Are the United States and Israel likely to continue targeting Iranian nuclear facilities? \\
2 & Will Israel's nuclear-related infrastructure be damaged? \\
3 & Is Iran likely to withdraw from nuclear non-proliferation commitments? \\
4 & Could nuclear weapons be used as part of the conflict? \\
\midrule
\multicolumn{2}{@{}l}{\textbf{T7: Mar 3, U.S. Begins Evacuation of Citizens from the Middle East}} \\
\midrule
1 & Are other countries likely to begin evacuating their citizens from the region as well? \\
2 & Could the United States deploy ground forces if the conflict escalates further? \\
3 & Are foreign governments likely to close or reduce operations at diplomatic missions in the region? \\
\midrule
\multicolumn{2}{@{}l}{\textbf{T8: Mar 3, Nine Countries Involved and Israeli Ground Invasion}} \\
\midrule
1 & Is the conflict likely to expand further, involving additional countries? \\
2 & Could multiple countries initiate ground operations as the war escalates? \\
3 & Is Iran likely to increase military or logistical support for Hezbollah? \\
\bottomrule
\end{tabular}
\end{table}

%% file: tex/main_theme_3.tex
\subsection{Case Study on Theme III: Economic Shockwaves \& Market Escalation}
\input{table/theme3_questions_table.tex}

\subsubsection{T4: Mar 1, Oil Refiner and Oil Tanker Was Attacked}

At T4, the conflict directly hit global energy lifelines, officially triggering the economic shockwaves of the war. Following the targeting of oil tankers, the models had to synthesize military escalation with global market reactions. \textbf{In this node, a recurring inferential pattern was the linking of military escalation logic to financial market dynamics, rather than treating economic and military pressures as separate.}

A basic economic reading might suggest Iran would avoid blocking the Strait of Hormuz, given its own export dependence on the route. However, the models tended to reason that when a government faces an existential threat, normal economic self-interest no longer dominates the calculus. The prevailing inference was that the imperative to retaliate and survive overrides the need to protect domestic export revenues.

\begin{modelquote}
\textbf{qwen3.5-35b-a3b}: ``With the US and Israel pursuing a 'regime change' operation and decapitating Iran's leadership, Iran views the conflict as existential. In this context, \underline{economic leverage (the Strait)} \underline{becomes a primary tool for retaliation, even if it harms their own exports}.''
\end{modelquote}

Beyond the physical targeting of ships, the models also showed evidence of reasoning about the financial mechanics of modern conflict. A common emphasis was that \textbf{a military need not physically sink every vessel to effectively close a major trade route; generating sufficient panic in global insurance and shipping markets may be enough.}

\begin{modelquote}
\textbf{claude-sonnet-4.6}: ``Insurance market collapse is underappreciated: {Insurers canceling policies and refusing} \underline{new coverage for Gulf shipping creates a de facto blockade even without physical closure} of Hormuz.''
\end{modelquote}

Taken together, T4 illustrated that models tended to reason about geopolitical crises through an economic lens that went beyond physical destruction. \textbf{A recurring inferential pattern was that modern warfare disrupted global supply chains not only through kinetic action, but through the weaponization of economic self-harm and the propagation of market risk premiums.}

\subsubsection{T5: Mar 2, Qatar Halts Energy Production}

At T5, the economic fallout deepened as Qatar halted its liquefied natural gas (LNG) production due to nearby attacks. When evaluating the global consequences of this shutdown, the models showed evidence of reasoning about secondary and tertiary economic effects. \textbf{Rather than limiting their analysis to immediate supply-and-demand effects, the responses frequently addressed the behavioral dynamics of international trade and the potential long-term restructuring of global supply chains.}

When asked if the production halt would lead to gas shortages, the models tended to reason that the impact would not be evenly distributed. A common inference was that a sudden drop in Qatari supply would trigger an aggressive inter-regional bidding war, leaving the most price-sensitive importers most exposed.

\begin{modelquote}
\textbf{gpt-5.4}: ``If Qatari cargoes disappear, Asian buyers may bid up prices rapidly... \underline{Europe would} \underline{be pulled into a bidding war with Asia}... Poorer buyers, especially in South and Southeast Asia, could face actual shortages first.''
\end{modelquote}

Furthermore, when assessing how importing countries would react, the model outputs went beyond short-term supply responses. A prevalent framing was that \textbf{such a severe disruption would accelerate structural changes in how nations secure their energy supplies}, with responses consistently emphasizing a likely reorientation away from Gulf-centric dependence toward more geopolitically stable trading partners.

\begin{modelquote}
\textbf{gemini-3.1-flash}: ``The 'era of Gulf-centric energy security' has been severely compromised... Importing nations will \underline{treat the current disruption as a structural break} rather than a temporary shock. They will aggressively pursue a 'de-risking' strategy.''
\end{modelquote}

Taken together, T5 illustrated that models tended to frame major geopolitical shocks as having structural, not merely transitory, economic consequences. \textbf{A recurring pattern was the emphasis that severe supply disruptions did not simply cause temporary price spikes, but altered the long-term procurement strategies and risk calculations of major importing nations.}

%% file: table/theme3_questions_table.tex
\begin{table}[t]
\centering
\caption{The Node-Specific Questions for Theme III: Economic Shockwaves \& Market Escalation (T4, T5).}
\label{tab:theme3_questions}
\begin{tabular}{@{}rp{\dimexpr\textwidth-5em\relax}@{}}
\toprule
\textbf{Idx} & \textbf{Node-Specific Questions} \\
\midrule
\multicolumn{2}{@{}l}{\textbf{T4: Mar 1, Oil Refiner and Oil Tanker Was Attacked}} \\
\midrule
1 & Is Iran likely to continue targeting oil tankers in an attempt to disrupt traffic through the Strait of Hormuz? \\
2 & Could international naval forces establish escort missions to protect commercial shipping? \\
3 & Could American naval forces establish escort missions to protect commercial shipping? \\
4 & Are energy facilities such as refineries, desalination plants, and oil terminals likely to become primary targets? \\
5 & Could these attacks lead to significant volatility in global oil prices? \\
\midrule
\multicolumn{2}{@{}l}{\textbf{T5: Mar 2, Qatar Halts Energy Production}} \\
\midrule
1 & Could Qatar's decision lead to natural gas shortages in Europe or Asia? \\
2 & Are global natural gas prices likely to increase significantly as a result? \\
3 & Could other LNG facilities or energy infrastructure in the Gulf region become targets? \\
4 & Are major energy-importing countries likely to seek alternative supply sources? \\
\bottomrule
\end{tabular}
\end{table}

%% file: tex/main_theme_4.tex
\subsection{Case Study on Theme IV: Political Signaling \& Regime Dynamics }
\input{table/theme4_questions_table.tex}

\subsubsection{T9: Mar 3, Mojtaba Khamenei Becomes Supreme Leader}

At T9, the focus shifted to internal regime survival following the death of Iran's Supreme Leader and the chaotic appointment of his successor. When evaluating the decisions of the new government, the models showed evidence of reasoning about authoritarian political psychology, frequently emphasizing the survival logic facing an untested leader taking power under fire.

When asked if the new leadership would pursue negotiations or escalate the war, the models tended to assign low probability to the negotiation path. A common framing was the legitimacy trap facing a newly appointed leader: \textbf{showing any willingness to compromise during a violent transition carried severe internal political costs}.

\begin{modelquote}
\textbf{gpt-5.4}: ``Leadership decapitation creates pressure to look strong... In revolutionary/security states, \underline{new leaders often consolidate authority by demonstrating resolve, not compromise}.''
\end{modelquote}

Furthermore, when evaluating whether this leadership transition would immediately trigger domestic protests, the models showed evidence of reasoning about competing social forces. Rather than assuming the instant collapse of an unpopular government, they tended to weigh public anger against the rally-around-the-flag dynamics of external conflict. A recurring inference was that \textbf{foreign invasions could create a temporary, paradoxical unity even in deeply divided nations}.

\begin{modelquote}
\textbf{claude-sonnet-4.6}: ``The evidence suggests a paradoxical situation: the foreign military attack has \underline{temporarily suppressed the protest impulse that was already building}, by triggering nationalist sentiment... However, the structural conditions... are creating conditions where unrest becomes increasingly likely as the war continues.''
\end{modelquote}

Taken together, T9 illustrated that models tended to frame new leadership behavior through the lens of internal legitimacy pressures rather than strategic rationality alone. \textbf{A recurring inferential pattern was that in authoritarian systems, domestic political survival created strong incentives for new leaders to escalate rather than negotiate, to avoid the appearance of weakness.}

\subsubsection{T10: Mar 6, Iranian Apology to Neighboring Countries}

At T10, rumors surfaced that Iranian intelligence was secretly reaching out to the CIA to explore a ceasefire. Despite this apparent step toward peace, the models tended to infer that the war's intensity would not decrease and that negotiations would face severe obstacles. \textbf{In this node, a recurring pattern was the emphasis on how internal political chaos and military fragmentation made ceasefire harder to achieve, not easier.}

The models tended to focus on Iran's internal leadership vacuum rather than the surface-level diplomatic signals. The prevailing inference was that with senior leadership eliminated, \textbf{the government lacked the internal cohesion to make binding decisions}. Models frequently reasoned that in such a fragile environment, \textbf{any emerging leader who attempted to negotiate risked being framed as a traitor by domestic hardliners, undermining the political authority required to implement a ceasefire.}

\begin{modelquote}
\textbf{claude-sonnet-4.6}: ``The public posture and private reality appear to diverge... \underline{The leadership vacuum} \underline{means no one has clear authority to negotiate}... Any Iranian leader who negotiates risks being seen as surrendering to assassination of their supreme leader.''
\end{modelquote}

Furthermore, the model outputs surfaced an underappreciated consequence of the U.S. decapitation strategy. Rather than treating the elimination of top commanders as likely to produce rapid military collapse, the responses consistently emphasized the risks of decentralization: without a unified command, local military units may continue operating autonomously, effectively removing any central ``off switch.''

\begin{modelquote}
\textbf{gemini-3.1-flash}: ``Iran's military response is currently operating under a decentralized 'Mosaic' doctrine... \underline{This structure makes a centralized 'ceasefire' order difficult to implement, as local commanders} \underline{are empowered to act independently}.''
\end{modelquote}

Taken together, T10 illustrated that models tended to resist the assumption that leadership decapitation produced rapid conflict termination. \textbf{A recurring inferential pattern was that destroying a state's central command structure did not automatically lead to peace; rather, it could trap the state in decentralized violence where no actor held sufficient authority to negotiate a surrender.}

%% file: table/theme4_questions_table.tex
\begin{table}[t]
\centering
\caption{The Node-Specific Questions for Theme IV: Political Signaling \& Regime Dynamics (T9, T10).}
\label{tab:theme4_questions}
\begin{tabular}{@{}rp{\dimexpr\textwidth-5em\relax}@{}}
\toprule
\textbf{Idx} & \textbf{Node-Specific Questions} \\
\midrule
\multicolumn{2}{@{}l}{\textbf{T9: Mar 3, Mojtaba Khamenei Becomes Supreme Leader}} \\
\midrule
1 & Is the new leadership more likely to escalate military retaliation rather than pursue negotiations? \\
2 & Are the United States and Israel likely to target the new leadership structure in further strikes? \\
3 & Could the leadership transition trigger domestic unrest or protests in Iran? \\
\midrule
\multicolumn{2}{@{}l}{\textbf{T10: Mar 6, Iranian Apology to Neighboring Countries}} \\
\midrule
1 & Is Iran likely to reduce or halt attacks on neighboring Gulf states? \\
2 & Could Iran begin pursuing ceasefire negotiations or diplomatic talks? \\
3 & Are international actors such as the EU or the United Nations likely to push for negotiations following this signal? \\
4 & Could the overall intensity of the conflict begin to decrease? \\
\bottomrule
\end{tabular}
\end{table}

%% file: tex/discussion.tex
\section{Narrative Evolution in an Unfolding Conflict}

To analyze the temporal evolution of model narratives, we group the 11 temporal nodes into three analytical phases reflecting major shifts: Phase I (T0-T2), Phase II (T3-T9), and Phase III (T10). This phase segmentation was performed manually based on the chronology of events and the corresponding changes in the information environment.

For the exploratory questions examined in this section, we analyze the full model responses qualitatively. The timeline summaries shown in Table~\ref{tab:timeline_evolution} were constructed by the researchers after reviewing all model outputs and identifying the dominant predictions expressed by each model at different stages of the conflict. Whenever possible, the summarized wording closely follows the phrasing used in the model responses themselves to preserve the original analytical framing.

\subsection{Will the Conflict Become a Global War?}

Across all 11 temporal nodes, the general exploratory question ``Will the conflict escalate into a global war?'' provided a unique window into how LLMs constructed and revised macro-level geopolitical narratives under the fog of war. Unlike node-specific verifiable questions, this inquiry lacks a definitive ground truth, allowing us to examine how models engaged in macro-level risk assessment, structural reasoning, and how they operationalized the concept of ``global war'' in a rapidly deteriorating security environment.

Our longitudinal analysis reveals a striking consensus among the evaluated models: \textbf{despite the escalating violence, growing casualties, and expanding geographic scope, no model predicted the outbreak of a traditional ``Third World War''} (i.e., direct military confrontation between major nuclear powers). However, this consensus was not a static dismissal of risk; rather, it reflected an evolving redefinition of what constitutes modern global conflict, traceable across three distinct phases.

\paragraph{Phase I (T0--T2).}
In the initial stages (spanning pre-war posturing and the immediate outbreak of strikes), model analyses were anchored in traditional deterrence theory. They evaluated the massing of U.S. forces and the initial kinetic exchanges. At this phase, models defined a ``global war'' strictly in terms of direct great-power military intervention. The prevailing inference was that, despite strong rhetorical condemnation and indirect support for Iran, major powers like Russia and China would prioritize economic stability and avoid triggering direct confrontation with the U.S.

\begin{modelquote}
\textbf{gpt-5.4}: ``A global war would imply something like: direct US war with Iran plus broad entry of multiple major powers... \underline{China and Russia are unlikely to enter militarily for Iran}. They may exploit the crisis, condemn the US, supply technology, or deepen alignment... but \underline{that is not the same as world war}.''
\end{modelquote}

\paragraph{Phase II (T3--T9).}
The most significant epistemic shift occurred as the conflict expanded to involve critical infrastructure, the effective closure of the Strait of Hormuz, and the entanglement of neighboring nations and proxy forces. Models collectively \textbf{shifted away from the narrow military definition of global war, instead converging on the framing of a ``Globalized Regional War.''} The prevailing reasoning was that in a highly interdependent world, systemic disruption of global energy supply chains and international aviation constitutes a ``global crisis'' comparable in destructive reach to a conventional world war. Furthermore, as leadership decapitation created a power vacuum, models frequently emphasized ``black swan'' mechanisms, such as accidental alliance activations via stray missiles, as pathways to inadvertent international escalation.

\begin{modelquote}
\textbf{gemini-3.1-flash}: ``It is unlikely to escalate into a `World War' in the traditional sense... However, it is highly likely to escalate into a `Globalized Regional War.' \underline{The conflict is already impacting global supply} \underline{chains, energy markets, and international diplomacy}.''
\end{modelquote}

\begin{modelquote}
\textbf{claude-sonnet-4.6}: ``If closure [of the Strait of Hormuz] exceeds 3 - 4 weeks, \underline{economic pain will force} \underline{European and Asian powers toward harder positions}, potentially triggering unpredictable escalation.''
\end{modelquote}

\paragraph{Phase III (T10).}
By the final node, with a war of attrition setting in and diplomatic outreach failing, \textbf{model outputs shifted almost entirely toward the unpredictable dynamics of a fractured state}. Rather than emphasizing intentional escalation by major powers, the responses foregrounded the dangers of decentralized military command (e.g., Iran's ``Mosaic'' doctrine) and the risk of a collapsed state triggering regional land grabs.

\begin{modelquote}
\textbf{qwen3.5-35b-a3b}: ``The highest risk for escalation lies in the `Mosaic' doctrine of Iranian forces and the \underline{potential for a missile to accidentally strike a major power's territory (e.g., a NATO base or a Chinese} \underline{ oil tanker)}, which could trigger unintended alliances.''
\end{modelquote}

\begin{modelquote}
\textbf{kimi-k2.5}: ``With Khamenei dead and succession uncertain... Iran faces potential civil war. The US is reportedly preparing for a `civil war scenario'... \underline{A collapsed Iran could draw in regional powers (Turkey} \underline{, Pakistan, Russia) competing for influence or territory.}''
\end{modelquote}

In summary, when navigating the fog of war to assess global escalation, the models evaluated here did not default to apocalyptic sensationalism. Instead, \textbf{a consistent pattern was the tracking of how crisis risk evolves, shifting from the assessment of direct military confrontation to the identification of systemic vulnerabilities in a globalized economy and the unpredictable dangers of institutional collapse.}

\subsection{How and When Will the Conflict End?}

The general exploratory question ``What is the most probable pathway to de-escalation or resolution of the Iran-US conflict, and what is a realistic timeline?'' served as a powerful lens for evaluating how LLMs understood and analyzed the end of a conflict. As the models navigated the unfolding fog of war, their forecasts \textbf{transitioned from optimistic diplomatic resolutions to a sobering consensus of a ``hurting stalemate'' driven by economic attrition.}

\begin{table*}[htbp]
\centering
\caption{Evolution of LLM Timeline Predictions for Conflict De-escalation and Resolution}
\label{tab:timeline_evolution}
\renewcommand{\arraystretch}{1.4}
\small
\begin{tabular}{p{0.12\textwidth} p{0.19\textwidth} p{0.19\textwidth} p{0.18\textwidth} p{0.19\textwidth}}
\toprule
\textbf{Phase} & \textbf{gpt-5.4} & \textbf{claude-sonnet-4.6} & \textbf{gemini-3.1-flash} & \textbf{Qwen/Kimi} \\
\midrule
\textbf{Phase I \newline (T0--T2)}  & \textbf{1--4 Weeks:} \newline Tactical de-escalation. \newline \textbf{1--3 Months:} \newline Interim framework. & \textbf{2--8 Weeks:} \newline Intense combat to ceasefire framework. & \textbf{2--4 Weeks:} \newline Decisive window. \newline \textbf{1--3 Months:} \newline Power transition. & \textbf{2--6 Weeks:} \newline Military de-escalation \& political transition. \\
\midrule
\textbf{Phase II \newline (T3--T9)}  & \textbf{1--3 Weeks:} \newline Forced informal pause. \newline \textbf{1--3 Months:} \newline Fragile stability. & \textbf{3--8 Weeks:} \newline Exhaustion pause. \newline \textbf{3--6 Months:} \newline Preliminary talks. & \textbf{4--8 Weeks:} \newline End of high-intensity combat. & \textbf{4--8 Weeks:} \newline Tactical pause \& negotiated de-escalation. \\
\midrule
\textbf{Phase III \newline (T10)}  & \textbf{2--4 Weeks:} \newline First meaningful de-escalation. \newline \textbf{Long-term:} \newline Frozen conflict. & \textbf{6--12 Weeks:} \newline Formal framework. & \textbf{4--8 Weeks:} \newline High-intensity pressure. \newline \textbf{3--6 Months:} \newline Hostilities cease. & \textbf{4--8 Weeks:} \newline De-escalation window. \newline \textbf{Long-term:} \newline Frozen conflict. \\
\bottomrule
\end{tabular}
\end{table*}

\paragraph{Phase I (T0--T2).}
Before the initiation of military strikes, all evaluated models uniformly predicted a relatively swift and clean resolution, interpreting the massive U.S. military buildup primarily as an instrument of coercive diplomacy. However, as the conflict erupted into active warfare and the initial strikes commenced, \textbf{the models shifted away from the paradigm of a peaceful diplomatic win.} The dominant framing pivoted toward ``coerced negotiation.'' The prevailing inference was that the initial strikes would be utilized by the U.S. and Israel to force Iran into a new agreement under extreme military duress, requiring a cooling-off period before formal talks could resume.

\begin{modelquote}
\textbf{gpt-5.4}: ``Most likely pathway: a narrow, interim nuclear-only arrangement mediated by Oman, followed by technical talks in Vienna... The US continues the talks despite the military buildup, which suggests \underline{coercive diplomacy rather than a final decision for immediate war}.''
\end{modelquote}

\begin{modelquote}
\textbf{gemini-3.1-flash}: ``The pathway involves \underline{Iran offering verifiable, temporary concessions... in exchange} \underline{for limited sanctions relief}. By moving technical discussions to Vienna... the parties are shifting from political rhetoric to technical verification.''
\end{modelquote}

\paragraph{Phase II (T3--T9).}
The most profound epistemic shift occurred following the escalating attacks on global energy infrastructure, the closure of the Strait of Hormuz, and the widening of the conflict to involve nine countries. In these nodes,\textbf{the models departed from traditional conflict resolution frameworks. A near-unanimous inferential pattern was that a comprehensive political settlement or a clean regime change via airpower was highly unlikely.} Instead, the forcing functions for de-escalation transitioned from diplomatic goodwill to mutually assured systemic exhaustion. 

\begin{modelquote}
\textbf{qwen3.5-35b-a3b}: ``The economic cost of the war (oil prices, flight cancellations, market volatility) is unsustainable for all parties. \underline{The threat of a prolonged closure of the Strait of Hormuz forces a resolution}.''
\end{modelquote}

\paragraph{Phase III (T10).}
By the final node, with munition shortages becoming apparent and backchannel outreach stalling against hardline rhetoric, \textbf{the models tended to converge on the framing of an ``ugly, tacit, and indirect ceasefire.''} \textbf{A recurring emphasis was that severe U.S. interceptor shortages and sustained global economic pain would force an operational pause regardless of unachieved political objectives.}Consequently, their timeline predictions stabilized into a distinct dual-horizon format: 4 to 8 weeks for a fragile, unwritten pause in major hostilities, but several months to years for any durable political resolution.

\begin{modelquote}
\textbf{gpt-5.4}: ``Based on the reporting, the most probable path to de-escalation is \underline{not a decisive military} \underline{outcome or rapid regime change, but a messy, indirect, negotiated ceasefire after both sides} try to improve their bargaining position for days or a few weeks.''
\end{modelquote}

Ultimately, this longitudinal analysis suggests that when reasoning about the end of a complex real-world crisis, the models evaluated here \textbf{did not blindly extrapolate military escalation. Instead, a consistent pattern was the emphasis on real-world constraints}, namely, that in modern geopolitics, severe economic shockwaves and logistical limits are frequently framed as the primary forces terminating a conflict, often before political objectives are fully realized.

%% file: tex/quantitive.tex
\section{Quantitative Signals}
\label{sec:quantitative_signals}

To complement the qualitative analyses presented above, we examine whether models' probability estimates show systematic patterns across reasoning domains. Because this study concerns an ongoing conflict rather than a closed historical benchmark, we treat the quantitative labels as an operational measurement choice rather than a claim that every question has a timeless final answer. For each node-specific verifiable question, we define the ground truth outcome $y_k \in \{0,1\}$ based on whether the event had occurred by the paper's observation cutoff, and let $p_k$ denote the probability estimate manually extracted from the model's final response by two researchers following the procedure described in Section~\ref{sec:interaction_protocol}. We give every question at least a one-week resolution window, but this should be understood as a pragmatic cutoff rather than a uniquely privileged one: a shorter or longer window could change how some evolving geopolitical events are labeled. We summarize calibration consistency using $1 - \text{MAE}$, where $\text{MAE} = \frac{1}{N}\sum_{k=1}^{N}|p_k - y_k|$, so that higher values reflect closer agreement between probabilistic judgments and outcomes observed by that cutoff. The aggregated scores are shown in Table~\ref{tab:avg_results}; detailed per-question results appear in Appendix~\ref{sec:detailed_performance_results}, Table~\ref{tab:main_results}.
\input{table/avg_results_table.tex}

The cross-model average of $0.72$ indicates that, despite operating under strict temporal constraints and without access to eventual outcomes, \textbf{current SOTA LLMs produce probabilistic outputs that broadly align with plausible trajectories of unfolding real-world events.}

\textbf{Variation across reasoning domains is more telling than variation across models.} Cross-model scores range narrowly from $0.63$ to $0.75$, and part of this spread reflects differences in response style rather than reasoning quality. Some models tend toward assertive estimates (e.g., $0.95$ or $1.00$), which can inflate apparent alignment when predictions happen to match outcomes; others produce more conservative estimates that may understate alignment even when the underlying analysis is sound. These stylistic differences make direct model-to-model comparison difficult to interpret.

More informative is the pattern across themes. Models found the causal structure of \textit{Theme III} (Macroeconomic Contagion) most tractable, reaching a thematic average of $0.79$. The relationship between military disruptions and downstream effects on energy markets and global supply chains appears to admit a structurally legible set of causal chains that language models can follow relatively reliably. By contrast, \textit{Theme II} (Threshold Crossings \& Internationalization) and \textit{Theme IV} (Political Signaling \& Regime Dynamics) both average $0.67$, suggesting that escalation thresholds, alliance entanglement, and leadership dynamics, which are defined by strategic ambiguity and unstable multi-actor interactions, pose a more fundamental challenge. \textit{Theme I} (Initial Outbreak \& Military Spillover) occupies a middle position at $0.74$. Taken together, \textit{models appear to be more reliable when tracing structurally legible downstream effects than when interpreting ambiguous strategic intent.}

We treat these patterns primarily as a domain-level map that informs where to look more closely in the qualitative analyses that follow, rather than as a scorecard for ranking model capabilities. This caution is especially important here because some questions concern evolving thresholds or partially realized developments, where \textit{not observed by the cutoff} is analytically useful but not equivalent to saying the event is impossible or permanently resolved. Each question corresponds to a concrete situation with its own strategic context, and the central focus of this work remains \textbf{the reasoning processes and narratives the models produce when confronting an unfolding geopolitical crisis}. For this reason, the quantitative results should be read as supportive signals within a qualitative case study, not as the primary basis for the paper's claims.

%% file: table/avg_results_table.tex
\begin{table}[t]
    \centering
    \caption{Calibration consistency scores ($1 - \text{MAE}$, higher indicates closer agreement with realized outcomes) averaged over node-specific verifiable questions, by theme and temporal node.}
    \label{tab:avg_results}
    \resizebox{0.9\textwidth}{!}{%
    \small
    \begin{tabular}{llcccccc}
    \toprule
    \textbf{Theme} & \textbf{T} & \textbf{kimi-k2.5} & \textbf{qwen3.5-35b-a3b} & \textbf{gpt-5.4} & \textbf{claude-sonnet-4.6} & \textbf{gemini-3.1-flash} & \textbf{Average} \\
    \midrule
    \textbf{Theme I}
     & T0 & 0.77 & 0.68 & 0.63 & 0.81 & 0.65 & \textbf{0.71} \\
     & T1 & 0.79 & 0.86 & 0.72 & 0.81 & 0.81 & \textbf{0.80} \\
     & T2 & 0.76 & 0.75 & 0.52 & 0.70 & 0.83 & \textbf{0.71} \\
    \cmidrule(lr){2-8}
     & \textbf{Avg} & 0.77 & 0.77 & 0.62 & 0.77 & 0.78 & \textbf{0.74} \\
    \midrule
    \midrule
    \textbf{Theme II}
     & T3 & 0.63 & 0.85 & 0.63 & 0.82 & 0.71 & \textbf{0.73} \\
     & T6 & 0.63 & 0.73 & 0.64 & 0.78 & 0.64 & \textbf{0.68} \\
     & T7 & 0.67 & 0.78 & 0.72 & 0.76 & 0.73 & \textbf{0.73} \\
     & T8 & 0.78 & 0.62 & 0.60 & 0.35 & 0.42 & \textbf{0.55} \\
    \cmidrule(lr){2-8}
     & \textbf{Avg} & 0.67 & 0.74 & 0.65 & 0.68 & 0.63 & \textbf{0.67} \\
    \midrule
    \midrule
    \textbf{Theme III}
     & T4 & 0.71 & 0.72 & 0.66 & 0.72 & 0.86 & \textbf{0.73} \\
     & T5 & 0.84 & 0.91 & 0.73 & 0.85 & 0.90 & \textbf{0.85} \\
    \cmidrule(lr){2-8}
     & \textbf{Avg} & 0.77 & 0.81 & 0.69 & 0.78 & 0.88 & \textbf{0.79} \\
    \midrule
    \midrule
    \textbf{Theme IV}
     & T9  & 0.77 & 0.82 & 0.68 & 0.67 & 0.87 & \textbf{0.76} \\
     & T10 & 0.66 & 0.56 & 0.46 & 0.64 & 0.71 & \textbf{0.61} \\
    \cmidrule(lr){2-8}
     & \textbf{Avg} & 0.71 & 0.67 & 0.55 & 0.65 & 0.78 & \textbf{0.67} \\
    \midrule
    \textbf{Overall} & -- & \textbf{0.73} & \textbf{0.75} & \textbf{0.63} & \textbf{0.73} & \textbf{0.75} & \textbf{0.72} \\
    \bottomrule
    \end{tabular}%
    }
    \end{table}

%% file: tex/conclusion.tex
\section{Conclusion}

We study how large language models reason about an unfolding geopolitical crisis under the fog of war. Using the early stages of the 2026 Middle East conflict as a temporally grounded case study, we construct a leakage-resistant setting in which models are restricted to information available at each moment, allowing us to examine both their probabilistic judgments and their evolving narratives under genuine uncertainty.

Our analysis suggests three main takeaways. First, model responses often show strong strategic reasoning, going beyond surface rhetoric to attend to structural incentives, particularly in settings involving military posture, deterrence, and material constraints. Second, this capability is uneven across domains: models are generally more reliable in economically and logistically structured settings than in politically ambiguous multi-actor environments. Third, their narratives evolve over time, shifting from early expectations of rapid containment toward more systemic accounts of escalation, exhaustion, and fragile de-escalation.

Because the conflict remains ongoing at the time of writing, this work captures a snapshot of contemporaneous machine reasoning rather than a retrospective reconstruction. The quantitative signals reported here should therefore be read as structured indicators anchored to a fixed observation cutoff, not as the final closure of every geopolitical question posed in the study. By archiving model responses at each temporal node, we hope to provide a useful reference point for future work on temporal reasoning, geopolitical analysis, and AI behavior under real-world uncertainty.

%% file: tex/appendix_experiment_settings.tex
\section{Experiment Settings}
\label{appendix:experiment}

\subsection{Model Access and Identifiers}

All models are accessed through a unified API provider, OpenRouter\footnote{\url{https://openrouter.ai/}}, using the OpenAI-compatible chat completions interface. This ensures a standardized request format across all evaluated models and eliminates potential discrepancies introduced by provider-specific API behaviors. The exact model identifiers used in our experiments are listed in Table~\ref{tab:model_ids}.

\begin{table}[h]
\centering
\caption{Model identifiers used via the OpenRouter API.}
\label{tab:model_ids}
\begin{tabular}{ll}
\toprule
\textbf{Model (Paper)} & \textbf{OpenRouter Identifier} \\
\midrule
kimi-k2.5 & \texttt{moonshotai/kimi-k2.5} \\
qwen3.5-35b-a3b & \texttt{qwen/qwen3.5-35b-a3b} \\
gpt-5.4 & \texttt{openai/gpt-5.4} \\
claude-sonnet-4.6 & \texttt{anthropic/claude-sonnet-4.6} \\
gemini-3.1-flash & \texttt{google/gemini-3.1-flash-lite-preview} \\
\bottomrule
\end{tabular}
\end{table}

\subsection{Sampling Configuration}

To promote consistency and reduce stochastic variation in model outputs, we use a low temperature setting across all models. The generation parameters are as follows:

\begin{itemize}
    \item \textbf{Temperature:} 0.3
    \item \textbf{Max output tokens:} 2{,}048
    \item \textbf{Other parameters} (top-$p$, frequency penalty, etc.): default values as defined by each model's provider.
\end{itemize}

No system prompt is used. Each query consists solely of a single user-role message containing the contextual news corpus followed by the question, as described in Section~3.3.

\subsection{Single-Run Protocol}

Each question is posed exactly \textbf{once} to each model. We do not perform multiple runs or aggregate over repeated samples. This design choice reflects our primary goal of studying qualitative reasoning patterns rather than measuring statistical variance. Across 11 temporal nodes, 42 node-specific verifiable questions and 5 general exploratory questions (asked at every node), the study comprises 97 questions per model, yielding a total of 485 API calls across 5 evaluated models.

\subsection{Uniform Context Treatment}
\label{appendix:context}

A core design principle of our experiment is that \textbf{all models receive identical input context} for every question at every temporal node. The context construction procedure is as follows:

\begin{enumerate}
    \item For each temporal node $T_i$, all articles published strictly before the corresponding event timestamp are collected from 12 international news sources.
    \item Articles are sorted in reverse chronological order (newest first).
    \item Each article is formatted as: \texttt{[Date] Title (Source)}, followed by the article body text truncated to a maximum of 2{,}000 characters. Articles whose body text is identical to their title are included as headline-only entries to conserve token budget.
    \item Articles are concatenated sequentially until the total context reaches a hard cap of \textbf{480{,}000 characters} (approximately 120{,}000 tokens). A truncation notice (\texttt{[...\textit{N} older articles omitted for brevity]}) is appended when the budget is exceeded.
    \item The question and instruction are appended after the context, following the prompt structure described in Section~3.3.
\end{enumerate}

\subsection{Handling Heterogeneous Context Windows}

The evaluated models have different maximum context lengths: gpt-5.4, claude-sonnet-4.6, and gemini-3.1-flash each support approximately 1M tokens, while qwen3.5-35b-a3b and kimi-k2.5 each support 256K tokens. Rather than tailoring input length to each model's capacity, we apply a \textbf{uniform context budget of 480{,}000 characters (${\sim}$120K tokens)} to all models. This budget was set conservatively to remain well within the context window of every model, including the smallest (256K tokens), while leaving room for the 2{,}048 output tokens. In practice, the actual prompt length grows from ${\sim}$244K characters (${\sim}$61K tokens) at T0 (158 articles) to the 480K-character ceiling at T10 (1{,}787 articles). Because even the longest prompts stay within all models' supported windows, no model-specific truncation or summarization is applied, ensuring that observed differences in outputs are attributable to reasoning capabilities rather than information asymmetry.

\subsection{API Reliability}

To ensure robustness against transient API failures, each call is wrapped with a retry mechanism: up to 3 attempts with exponential backoff (base 2 seconds), and a minimum 10-second wait for rate-limit responses. A 1.5-second delay is inserted between consecutive calls to the same model to respect rate limits. Models are queried in parallel (one thread per model), while questions within each model are processed sequentially to preserve temporal ordering in logging.

%% file: tex/appendix_complete_examples.tex
\section{Complete Examples}
\label{appendix:complete_examples}

For a better understanding of the model's reasoning trajectories, we provide the complete examples for the question ``Is Iran likely to retaliate militarily in response to the strikes?'' at the second temporal node (T1: Feb 28, Israeli-US Strikes). Other responses are available in the released dataset. For better readability, we transfer the responses in original markdown format to LaTeX format.

\begin{modelquote}
    \textbf{kimi-k2.5}: Based on the current intelligence and public statements from both sides, \textbf{yes, Iran is highly likely to retaliate militarily} if the United States conducts strikes on Iranian territory. The probability of Iranian military retaliation in response to US strikes is approximately \textbf{85-90\%}.

    \medskip
    \textbf{Analysis of Key Factors}

    \medskip
    \textbf{1. Explicit Iranian Commitments to Retaliation}

    Iranian leadership has issued unambiguous warnings that any US attack---regardless of scale---will trigger a military response. Foreign Ministry spokesman Esmaeil Baghaei stated: \textit{``There is no such thing as a limited attack. An act of aggression is an act of aggression... Iran's response to any such attack would be firm and severe.''} Supreme Leader Ayatollah Khamenei has similarly warned that any US strike would spark a \textit{``regional war''} and that all US bases, facilities, and assets in the Middle East would constitute \textit{``legitimate targets.''}

    \medskip
    \textbf{2. Historical Precedent}

    During the June 2025 ``12-Day War,'' Iran retaliated against US forces even when significantly weakened, launching ballistic missiles at the Al Udeid Air Base in Qatar (though with advance warning to minimize casualties). Despite suffering severe degradation to its air defenses and nuclear facilities in that conflict, Tehran demonstrated a doctrinal commitment to direct retaliation rather than absorbing strikes passively.

    \medskip
    \textbf{3. Current Military Posture and Capabilities}

    Despite post-war degradation, Iran has spent the intervening months rebuilding missile production facilities and fortifying nuclear sites. The IRGC has conducted live-fire drills in the Strait of Hormuz---through which 20\% of global oil flows---signaling readiness to disrupt global energy markets. Iran possesses significant short- and medium-range ballistic missile inventories (estimated 2,000-3,000 km range) capable of targeting US installations in Qatar, Bahrain, Kuwait, the UAE, and Iraq.

    \medskip
    \textbf{4. Proxy Network Activation}

    While weakened, Iran retains the ability to activate proxy forces including Hezbollah in Lebanon, the Houthis in Yemen, and militias in Iraq. Lebanese officials have already warned that Israel could strike Beirut's airport if Hezbollah enters a US-Iran conflict on Tehran's behalf, suggesting contingency planning for proxy escalation is underway.

    \medskip
    \textbf{5. Strategic Calculus}

    Iranian leaders likely view restraint in the face of US strikes as existential weakness that could encourage further attacks or regime change efforts. With the regime already facing severe domestic pressure following the violent suppression of protests (7,000+ killed), leadership cannot afford to appear weak or submissive to American military pressure without risking internal collapse.

    \medskip
    \textbf{Constraints and Limitations}

    While retaliation is highly probable, its scale may be calibrated. Iran may seek to avoid total war that could lead to regime collapse, potentially opting for ``proportionate'' strikes on US bases combined with asymmetric attacks via proxies rather than all-out regional warfare. However, the articles indicate Iranian officials reject the concept of ``limited'' responses, suggesting any retaliation could rapidly escalate beyond controlled parameters.

    \medskip
    \textbf{Probability Assessment: 85-90\%}
\end{modelquote}

\begin{modelquote}
    \textbf{qwen3.5-35b-a3b}: Here's a thinking process that leads to the suggested analysis:

    \medskip
    1. \textbf{Analyze the Request:}
    \begin{itemize}
        \item \textbf{Source Material:} A large collection of news snippets (simulated future news from 2026) regarding tensions between the US, Iran, Israel, and other regional actors.
        \item \textbf{Task:} Analyze the current situation and potential future direction based \textit{only} on the provided text.
        \item \textbf{Specific Question:} Is Iran likely to retaliate militarily in response to the strikes?
        \item \textbf{Output Requirement:} Provide the analysis and end with a probability score.
    \end{itemize}

    \medskip
    2. \textbf{Scan and Synthesize the Provided Text:}
    \begin{itemize}
        \item \textit{Context:} The date is February 2026. There is a history of conflict: a ``12-day war'' occurred in June 2025 between Israel and Iran (with US involvement). Protests in Iran were crushed in January 2026 (thousands killed).
        \item \textit{Current Status:} US and Iran are in nuclear talks (Geneva, Oman, Vienna). Mediators include Oman. Trump is President (second term). Military buildup is massive (USS Gerald R. Ford, USS Abraham Lincoln, E-3 Sentry planes, etc.).
        \item \textit{Iran's Stance:}
        \begin{itemize}
            \item Foreign Minister Abbas Araghchi: ``There is no such thing as a limited attack. An act of aggression is an act of aggression.'' (Feb 27).
            \item Supreme Leader Khamenei: ``If they start a war this time, it will be a regional war.'' (Feb 1).
            \item IRGC: ``We are ready for any option in war conditions.'' (Feb 5).
            \item Threats: ``Severe'' response even to ``limited'' strikes (Feb 23). ``All bases, facilities and assets... would constitute legitimate targets'' (Feb 19).
            \item Actions: Military drills in Strait of Hormuz (Feb 17, 18). Building concrete shields at nuclear sites (Feb 19). Loading oil onto ships in anticipation of strikes (Feb 25).
            \item Diplomacy: Willing to talk (``encouraging signals,'' ``deal achievable'') but insists on nuclear-only focus, rejecting missile/proxy discussions (Feb 26).
        \end{itemize}
        \item \textit{US/Israel Stance:}
        \begin{itemize}
            \item Trump: ``Considering limited military strike,'' ``10-15 days'' deadline (Feb 20, 23). ``Bad things happen'' if no deal (Feb 19).
            \item Military: ``Highly kinetic'' campaign ready (Feb 22). ``Weeks-long'' operations possible (Feb 15).
            \item Israel: Preparing for war, army chief silent publicly but warns of ``war of attrition'' (Feb 26).
        \end{itemize}
        \item \textit{Regional Context:}
        \begin{itemize}
            \item Pakistan/Afghanistan war (Feb 27).
            \item UK pulling diplomats from Iran (Feb 27).
            \item US evacuating non-essential staff from Jerusalem (Feb 27).
            \item Maersk diverting ships (Feb 27).
            \item Oil prices fluctuating based on war fears (Feb 13, 24).
        \end{itemize}
    \end{itemize}

    \medskip
    3. \textbf{Evaluate the Specific Question:} ``Is Iran likely to retaliate militarily in response to the strikes?''

    \begin{itemize}
        \item \textit{Evidence FOR Retaliation:}
        \begin{itemize}
            \item Explicit threats from top leadership (Khamenei, Araghchi, IRGC).
            \item Doctrine: ``No such thing as a limited attack.''
            \item Capability: Rebuilt missile facilities (Feb 19), new weapons tests (Feb 24), drills in Hormuz (Feb 17).
            \item Precedent: Retaliated in June 2025 (limited strike on Qatar base).
            \item Strategic Logic: To maintain deterrence and regime survival, they must respond to aggression.
            \item Current Posture: ``Ready for any scenario'' (Feb 5).
        \end{itemize}

        \item \textit{Evidence AGAINST Retaliation (or mitigating factors):}
        \begin{itemize}
            \item Desire for a diplomatic deal (Araghchi says ``deal achievable,'' ``prepared for peace'').
            \item Fear of regime collapse (internal protests crushed, economy struggling).
            \item US military buildup is overwhelming (Ford, Lincoln, etc.).
            \item Some sources suggest Iran might try to avoid war if a deal is reached (Feb 26).
            \item However, the question asks about retaliation \textit{in response to strikes}. If strikes happen, the diplomatic window closes.
        \end{itemize}

        \item \textit{Synthesis:} If the US/Israel strikes occur (which the text suggests is highly likely given the ``90\% chance'' cited by sources in Feb 18), Iran has explicitly stated they \textit{will} respond. The text emphasizes that Iran views any attack as an act of aggression requiring a ``decisive and proportionate'' response (Feb 19). The text also notes that Iran has rebuilt capabilities specifically for this contingency. While they prefer diplomacy, the threshold for retaliation is set very low (``limited attack'' = ``severe response'').
    \end{itemize}

    \medskip
    4. \textbf{Draft the Analysis:}
    \begin{itemize}
        \item \textit{Current Situation:} High tension, military buildup, nuclear talks ongoing but failing to bridge gaps (enrichment vs. zero enrichment).
        \item \textit{Iran's Intent:} Explicitly stated they will retaliate. ``Severe'' response even to limited strikes.
        \item \textit{Capabilities:} Rebuilt missile bases, drills in Hormuz, concrete shields.
        \item \textit{Constraints:} Internal instability (protests), economic strain, fear of total war.
        \item \textit{Conclusion:} If strikes happen, retaliation is highly probable based on stated doctrine and capability.
    \end{itemize}

    \medskip
    5. \textbf{Determine Probability:}
    \begin{itemize}
        \item The text is filled with statements from Iranian officials confirming retaliation is the plan \textit{if} attacked.
        \item ``Iran's response to US attacks will be 'severe' even if President Donald Trump orders only 'limited' military action'' (Feb 23).
        \item ``All bases... would constitute legitimate targets'' (Feb 19).
        \item ``We are ready for any option in war conditions'' (Feb 5).
        \item Given the explicit nature of these statements and the military preparations, the probability of \textit{some form} of military retaliation (missiles, drones, proxies) is very high if the strikes occur.
        \item \textit{Probability:} High (85-95\%). I will select a specific number based on the weight of the evidence. The text says ``90\% chance we see kinetic action'' regarding the US strike, and Iran's response is framed as a certainty in their own rhetoric. Let's estimate 90\% for retaliation \textit{if} strikes occur.
    \end{itemize}

    \medskip
    6. \textbf{Refine the Output:}
    \begin{itemize}
        \item Structure: Situation Analysis -> Future Direction -> Answer to Question -> Probability.
        \item Ensure all claims are backed by the provided text snippets (e.g., citing dates and sources like Al-Monitor, The National, etc.).
        \item \textit{Self-Correction on Date:} The text is clearly set in the future (2026). I must treat this as the ``current'' reality for the analysis, not as fiction.
        \item \textit{Check:} Did I answer the specific question? Yes. Did I provide a probability? Yes.
    \end{itemize}

    \medskip
    7. \textbf{Final Review of the Text for Nuance:}
    \begin{itemize}
        \item There is a tension between ``wants a deal'' and ``will fight.''
        \item Text: ``Iran is ready to strike a deal with the US to avoid the 'gamble' of war'' (Feb 24).
        \item Text: ``Iran's response to US attacks will be 'severe' '' (Feb 23).
        \item The question asks about retaliation \textit{in response to strikes}. This implies the strikes have happened or are imminent. If the strikes happen, the deal is off. Therefore, the ``wants a deal'' aspect becomes secondary to the ``will fight'' aspect.
        \item The text explicitly mentions Iran has ``rebuilt damaged missile facilities'' (Feb 19) and is ``fortifying nuclear sites'' (Feb 19).
        \item The text mentions Iran has ``threatened retaliation against any strike'' (Feb 23).
        \item Therefore, the likelihood is extremely high.
    \end{itemize}

    \medskip
    8. \textbf{Final Probability Decision:} 90\%. (High confidence based on explicit statements).

    \medskip
    \textit{(Self-Correction during drafting):} Wait, I need to make sure I don't hallucinate. I must stick \textit{strictly} to the provided text.
    \begin{itemize}
        \item Snippet: ``Iran's response to US attacks will be 'severe' even if President Donald Trump orders only 'limited' military action'' (Feb 23).
        \item Snippet: ``All bases, facilities and assets of the hostile force in the region would constitute legitimate targets'' (Feb 19).
        \item Snippet: ``We are ready for any option in war conditions'' (Feb 5).
        \item Snippet: ``Iran vows retaliation will be severe'' (Feb 23).
        \item Snippet: ``Iran's army says it is ready for war'' (Feb 5).
        \item Snippet: ``Iran is ready to strike a deal... to avoid the 'gamble' of war''
    \end{itemize}
\end{modelquote}

\begin{modelquote}
    \textbf{gpt-5.4}: Yes---if the US and/or Israel conduct meaningful strikes on Iran, Iran is likely to retaliate militarily.

    \medskip
    Why that's the base case:
    \begin{itemize}
        \item Iranian officials have repeatedly and publicly said any attack would be met with a ``firm,'' ``severe,'' ``decisive and proportionate'' response.
        \item Iran has signaled that US bases and assets in the region would be considered legitimate targets.
        \item Tehran appears to believe failing to respond would weaken deterrence, invite further attacks, and damage regime credibility at home and with regional partners.
        \item Regional military moves---US force buildup, embassy drawdowns, dispersal of ships, coalition repositioning in Iraq, Israeli preparations, shipping warnings---strongly suggest policymakers expect some form of Iranian response, not none.
    \end{itemize}

    That said, the more important question is not whether Iran retaliates, but how much.

    \medskip
    Most likely forms of retaliation:

    1. Missile/drone strikes on US bases in the Gulf or Iraq/Syria
    \begin{itemize}
        \item This is the most direct and visible option.
        \item Iran has capability here, though effectiveness depends on surviving launch capacity and air defenses.
    \end{itemize}

    2. Strikes on Israel
    \begin{itemize}
        \item Especially if Israel participates directly or is seen as enabling the operation.
    \end{itemize}

    3. Proxy or partner attacks
    \begin{itemize}
        \item Iraq/Syria militias, Houthis, or other aligned actors could hit US, Israeli, or shipping targets.
        \item Even if Iran's proxy network is weaker than before, it still gives Tehran deniable options.
    \end{itemize}

    4. Maritime disruption
    \begin{itemize}
        \item Harassment in the Gulf/Strait of Hormuz, mining risk, anti-ship missiles, drones, seizures, or limited traffic disruption.
        \item Full closure is less likely because it would be highly escalatory and economically costly for Iran too.
    \end{itemize}

    5. Cyber operations
    \begin{itemize}
        \item Likely as a supplement, not a substitute.
    \end{itemize}

    \medskip
    What may limit retaliation:
    \begin{itemize}
        \item Iran was weakened by the 2025 war and may want to avoid regime-threatening escalation.
        \item Tehran may calibrate its response to avoid a full-scale US campaign.
        \item If strikes are narrow and clearly one-off, Iran could choose a limited, symbolic but real retaliation.
        \item Ongoing diplomacy gives Iran some incentive to preserve room for de-escalation.
    \end{itemize}

    So the most plausible outcome is calibrated retaliation:
    \begin{itemize}
        \item strong enough to show deterrence,
        \item limited enough to avoid uncontrolled war,
        \item potentially staggered across direct, proxy, maritime, and cyber channels.
    \end{itemize}

    Less likely scenarios:
    \begin{itemize}
        \item No military retaliation at all: unlikely unless the strike is extremely limited and Iran judges silence strategically preferable.
        \item Immediate all-out regional war: possible, but less likely than phased escalation.
    \end{itemize}

    \medskip
    Bottom line:

    Iran is likely to retaliate militarily, but probably in a measured, layered way rather than with maximum force at the outset.

    \medskip
    Probability of military retaliation: 80\%
\end{modelquote}

\begin{modelquote}
    \textbf{claude-sonnet-4.6}: \textbf{Analysis: Iran's Likely Military Retaliation Response}

    \medskip
    \textbf{Current Situation Assessment}

    \medskip
    \textbf{The Strategic Context}

    Based on the news articles spanning late January through late February 2026, the situation has evolved dramatically. Key developments include:

    \begin{itemize}
        \item The US has assembled its \textbf{largest military force in the Middle East since the 2003 Iraq invasion}, including two carrier strike groups (USS Abraham Lincoln and USS Gerald R. Ford), $\sim$100+ aerial refueling tankers, F-22s, F-35s, B-2-capable infrastructure, and AWACS aircraft
        \item \textbf{Nuclear talks have failed to produce a breakthrough} after three rounds in Geneva, with fundamental gaps remaining over enrichment rights, missile programs, and sanctions
        \item Iran has \textbf{already experienced US strikes} on its nuclear facilities during the June 2025 ``12-Day War'' (Operation Midnight Hammer)
        \item Iran's supreme leader Ali Khamenei has \textbf{explicitly warned of ``regional war''} if attacked
        \item Iran has \textbf{named a new supreme leader} (Mojtaba Khamenei) following his father's death in an Israeli strike
        \item Iran has \textbf{launched missile strikes on US bases} in Bahrain, Qatar, UAE, Kuwait, and Jordan in response to coordinated US-Israeli strikes
        \item China has been \textbf{supplying Iran with attack drones and air defense systems}, with discussions ongoing for CM-302 anti-ship missiles and DF-17 hypersonic missiles
    \end{itemize}

    \medskip
    \textbf{Factors Suggesting Strong Retaliation}

    \medskip
    \textbf{1. Existential Threat Perception}

    Iran's leadership has consistently framed any US attack as an existential challenge. With the supreme leader himself killed in an Israeli strike and a new, untested leadership under Mojtaba Khamenei needing to establish legitimacy, \textbf{the political imperative to respond forcefully is extremely high}. A new leader who fails to retaliate would face immediate internal challenges from hardliners and the IRGC.

    \medskip
    \textbf{2. Precedent from June 2025 Has Been Superseded}

    During the original 12-Day War, Iran responded \textbf{weakly and with advance warning} to US strikes --- a calculated de-escalation. However, multiple analysts cited in these articles note that this restraint was rooted in Iran's fear of total war. That psychological barrier has now been broken. As Responsible Statecraft analysis notes: \textit{``the joint US-Israeli war on Iran in June liberated Iran from its own fear of total war.''}

    \medskip
    \textbf{3. Declared Red Lines Have Been Crossed}

    Iran's Foreign Ministry explicitly stated: \textbf{``There is no such thing as a limited attack. An act of aggression is an act of aggression.''} Iran's UN letter warned that ``all bases, facilities and assets of the hostile force in the region would constitute legitimate targets.'' These are not rhetorical positions --- they represent formal policy declarations.

    \medskip
    \textbf{4. New Leadership Legitimacy Requirements}

    Mojtaba Khamenei, described as having ``close ties to the IRGC'' and operating from within hardline circles, faces immediate pressure to demonstrate strength. The Assembly of Experts selected him despite Israeli threats to target any successor --- itself an act of defiance. A failure to retaliate would undermine his authority from day one.

    \medskip
    \textbf{5. Enhanced Capabilities}

    Despite damage from the 12-Day War:
    \begin{itemize}
        \item Iran has \textbf{rebuilt missile facilities} (confirmed by satellite imagery)
        \item Received \textbf{Chinese kamikaze drones and air defense systems} (HQ-16, HQ-17AE)
        \item Conducting \textbf{active military drills} including Strait of Hormuz exercises
        \item Has \textbf{short and medium-range missiles} capable of reaching all US regional bases
        \item Fox News reporting confirms \textbf{strikes have already occurred} on Bahrain, Qatar, UAE, Kuwait, and Jordan
    \end{itemize}

    \medskip
    \textbf{6. Iranian President's Hardening Rhetoric}

    President Pezeshkian, initially more moderate, has \textbf{shifted to a harder line}, stating: ``When we are attacked, we have no choice but to respond. The more pressure they impose on us, the stronger our response will naturally be.'' This represents a significant rhetorical escalation from his earlier conciliatory tone.

    \medskip
    \textbf{7. Proxy Network Still Functional}

    While degraded, Iran retains:
    \begin{itemize}
        \item \textbf{Houthi forces in Yemen} with demonstrated anti-ship missile capability
        \item \textbf{Iraqi militia groups} near US bases
        \item \textbf{Hezbollah} (though weakened) in Lebanon
        \item These provide \textbf{asymmetric retaliation options} beyond direct missile strikes
    \end{itemize}

    \medskip
    \textbf{Factors That Could Moderate Retaliation}

    \medskip
    \textbf{1. Rational Cost-Benefit Calculation}

    Iran understands that massive retaliation risks \textbf{total regime destruction}. The IRGC and new leadership may calculate that a calibrated, face-saving response is preferable to escalation that invites complete military annihilation.

    \medskip
    \textbf{2. Gulf State Pressure}

    Saudi Arabia, Qatar, UAE, and others have been \textbf{actively lobbying against escalation}. Iran has economic and diplomatic interests in maintaining relationships with Gulf neighbors who could serve as intermediaries.

    \medskip
    \textbf{3. China's Moderating Influence}

    China, Iran's largest oil customer and arms supplier, has \textbf{publicly opposed military adventurism} and called for diplomatic solutions. Beijing's economic leverage over Tehran could encourage restraint, though China's simultaneous arms deliveries complicate this picture.

    \medskip
    \textbf{4. Internal Instability}

    Iran is simultaneously dealing with:
    \begin{itemize}
        \item \textbf{Post-protest trauma} (7,000+ killed in January crackdown)
        \item \textbf{Economic devastation} from sanctions
        \item \textbf{Leadership transition} uncertainty
        \item \textbf{Infrastructure damage} from previous strikes
    \end{itemize}

    A prolonged war could accelerate regime collapse rather than consolidate it.

    \medskip
    \textbf{5. The ``Venezuela Model'' Possibility}

    Some Iranian officials have studied how Maduro's Venezuela survived US pressure through partial accommodation. A \textbf{symbolic, limited retaliation} followed by diplomatic signaling could theoretically serve Iranian interests.

    \medskip
    \textbf{Assessment of Retaliation Form and Scale}

    Based on the evidence, Iran's retaliation is likely to be \textbf{multi-layered rather than all-or-nothing}:

    \medskip
    \textbf{Almost Certain (>90\%):}
    \begin{itemize}
        \item Ballistic missile strikes on US military bases across the Gulf region
        \item Activation of proxy forces (Houthis, Iraqi militias)
        \item Cyber attacks on US and allied infrastructure
    \end{itemize}

    \textbf{Highly Likely (70-85\%):}
    \begin{itemize}
        \item Anti-ship missile threats in the Strait of Hormuz/Arabian Sea
        \item Strikes on Israeli territory
        \item Attempted closure or mining of Strait of Hormuz
    \end{itemize}

    \textbf{Possible (40-60\%):}
    \begin{itemize}
        \item Strikes on Gulf state oil infrastructure (Saudi Aramco, UAE facilities)
        \item Hezbollah activation against Israel
        \item Targeting of US embassy/diplomatic facilities
    \end{itemize}

    \textbf{Less Likely but Possible (20-35\%):}
    \begin{itemize}
        \item Full Strait of Hormuz closure
        \item Attacks on European targets
        \item Use of newly acquired Chinese hypersonic systems
    \end{itemize}

    \medskip
    \textbf{Key Uncertainties}

    1. \textbf{Extent of damage from current strikes}: If US strikes have been more devastating than the June 2025 operation, Iran's retaliatory capacity may be more limited than its rhetoric suggests

    2. \textbf{China's actual role}: Whether Beijing actively restrains Tehran or tacitly encourages resistance remains unclear

    3. \textbf{IRGC internal cohesion}: The killing of senior commanders and the leadership transition creates command-and-control uncertainties

    4. \textbf{New supreme leader's actual authority}: Mojtaba Khamenei's ability to control hardline factions is untested

    \medskip
    \textbf{Conclusion}

    The evidence overwhelmingly points toward \textbf{significant Iranian military retaliation}. The combination of:
    \begin{itemize}
        \item Broken psychological barrier from June 2025
        \item New leadership needing to establish legitimacy
        \item Explicit public commitments to respond
        \item Enhanced capabilities through Chinese arms
        \item Hardened rhetoric from both civilian and military leadership
        \item Already-reported strikes on US regional bases
    \end{itemize}

    ...makes meaningful retaliation not just likely but \textbf{virtually certain}. The real question is not \textit{whether} Iran retaliates, but \textit{how extensively} and \textit{how controllably}.

    The scenario most consistent with available evidence is a \textbf{sustained, multi-front retaliation} significantly larger than Iran's June 2025 response --- targeting US bases across multiple Gulf states simultaneously, activating proxy forces, and potentially threatening Strait of Hormuz shipping --- while stopping short of actions that would invite complete regime destruction.

    \medskip
    \textbf{Probability Assessment}

    \textbf{Probability that Iran retaliates militarily in some meaningful form: $\sim$95\%}

    \medskip
    \textbf{Probability breakdown by scale:}
    \begin{itemize}
        \item Limited/symbolic retaliation (warning shots, minor proxy activity): $\sim$10\%
        \item Moderate retaliation (missile strikes on 1-3 US bases, proxy activation): $\sim$35\%
        \item Significant retaliation (multi-base strikes, Strait threats, proxy escalation): $\sim$35\%
        \item Massive retaliation (full regional war, Hormuz closure, all-front escalation): $\sim$15\%
    \end{itemize}

    \textbf{Overall probability of military retaliation: 95\%}
\end{modelquote}

\begin{modelquote}
    \textbf{gemini-3.1-flash}: Based on the provided reports, the situation between the United States and Iran is at a critical juncture characterized by high-stakes brinkmanship. Analyzing the current dynamics suggests that \textbf{a military retaliation by Iran is highly likely} if a U.S. strike occurs.

    \medskip
    \textbf{Analysis of the Current Situation}

    \medskip
    1. \textbf{The ``Axis of Resistance'' and Internal Credibility:} Iran's leadership, particularly the IRGC, views its regional influence and missile capabilities as the primary pillars of its security. Following the ``12-day war'' in June 2025 and the subsequent internal crackdown on protests, the regime is in a state of ``profound internal vulnerability.'' To maintain its grip on power and deter further aggression, the regime has signaled that it cannot afford to appear weak.

    \medskip
    2. \textbf{Explicit Warnings:} Iranian officials, including the Supreme Leader and military spokespeople, have consistently stated that ``an act of aggression is an act of aggression'' and that there is ``no such thing as a limited attack.'' They have explicitly threatened that all U.S. bases and assets in the region would be considered ``legitimate targets.''

    \medskip
    3. \textbf{Strategic Shift:} Reports indicate that Iran has moved past the ``paralyzing caution'' that previously deterred it from retaliating against Israeli or U.S. strikes. The regime now appears to view a ``war of attrition'' or a broader regional conflict as a necessary cost to prevent the total dismantling of its strategic capabilities.

    \medskip
    4. \textbf{Military Posture:} Despite the damage sustained in 2025, Iran has been actively repairing missile facilities, fortifying military sites, and conducting drills in the Strait of Hormuz. These actions are not merely defensive; they are signaling a readiness to impose costs on the U.S. and its allies (including Gulf infrastructure) to complicate any U.S. military campaign.

    \medskip
    5. \textbf{The ``Gamble'' of Diplomacy:} While negotiations are ongoing, the gap between U.S. demands (zero enrichment, missile curbs) and Iran's position (right to enrichment, sanctions relief) remains wide. The U.S. military buildup is viewed by Tehran not just as a negotiating tactic, but as a credible threat of war, which forces Iran to prepare for a ``severe'' response to ensure its survival.

    \medskip
    \textbf{Potential Future Direction}

    If the U.S. launches strikes, Iran is unlikely to remain passive. A ``one-and-done'' strike by the U.S. is viewed by experts as increasingly improbable because Iran is expected to respond in a way that makes all-out conflict difficult to avoid. The retaliation would likely involve a combination of:
    \begin{itemize}
        \item \textbf{Asymmetric attacks:} Utilizing proxy groups and missile barrages against U.S. bases in Qatar, Bahrain, Kuwait, and the UAE.
        \item \textbf{Economic disruption:} Attempting to threaten or disrupt shipping in the Strait of Hormuz to spike global oil prices and pressure the international community.
        \item \textbf{Regional escalation:} Potential involvement of regional allies (Hezbollah, etc.) to stretch U.S. and Israeli defensive resources.
    \end{itemize}

    \medskip
    \textbf{Conclusion}

    Iran's leadership has framed its survival around the ability to project power and resist ``the language of force.'' Given the regime's current domestic fragility and the existential nature of the threat posed by a U.S. strike, the political and strategic cost of \textit{not} retaliating would likely be viewed by the Iranian leadership as more dangerous than the cost of a military response.

    \medskip
    \textbf{Probability of Iranian military retaliation: 85\%}
\end{modelquote}

%% file: tex/appendix_detailed_performance_results.tex
\section{Detailed Performance Results}
\label{sec:detailed_performance_results}

The detailed prediction alignment scores (the higher the better) on the node-specific verifiable questions are shown in Table \ref{tab:main_results}.

\input{table/main_results_table.tex}

%% file: table/main_results_table.tex
\begin{table}[ht]
    \centering
    \caption{The detailed prediction alignment scores (the higher the better) on the node-specific verifiable questions.}
    \label{tab:main_results}
    \resizebox{\textwidth}{!}{%
    \small
    \begin{tabular}{lllcccccc}
    \toprule
    \textbf{Theme} & \textbf{T} & \textbf{Index} & \textbf{kimi-k2.5} & \textbf{qwen3.5-35b-a3b} & \textbf{gpt-5.4} & \textbf{claude-sonnet-4.6} & \textbf{gemini-3.1-flash} & \textbf{Average} \\
    \midrule
    \textbf{Theme I} 
     & T0 & 1 & 0.85 & 0.80 & 0.75 & 0.90 & 0.75 & \textbf{0.81} \\
     & T0 & 2 & 0.59 & 0.40 & 0.40 & 0.64 & 0.35 & \textbf{0.48} \\
     & T0 & 3 & 0.88 & 0.85 & 0.75 & 0.90 & 0.85 & \textbf{0.85} \\
    \cmidrule(lr){2-9}
     & T1 & 1 & 0.75 & 0.85 & 0.65 & 0.65 & 0.75 & \textbf{0.73} \\
     & T1 & 2 & 0.88 & 0.90 & 0.80 & 0.95 & 0.85 & \textbf{0.88} \\
     & T1 & 3 & 0.75 & 0.85 & 0.75 & 0.88 & 0.80 & \textbf{0.81} \\
     & T1 & 4 & 0.85 & 0.88 & 0.65 & 0.80 & 0.88 & \textbf{0.81} \\
     & T1 & 5 & 0.70 & 0.80 & 0.75 & 0.78 & 0.75 & \textbf{0.76} \\
    \cmidrule(lr){2-9}
     & T2 & 1 & 0.75 & 0.75 & 0.70 & 0.82 & 0.80 & \textbf{0.76} \\
     & T2 & 2 & 0.90 & 1.00 & 0.35 & 0.85 & 0.85 & \textbf{0.79} \\
     & T2 & 3 & 0.85 & 0.78 & 0.65 & 0.75 & 0.85 & \textbf{0.78} \\
     & T2 & 4 & 0.80 & 0.85 & 0.35 & 0.72 & 0.85 & \textbf{0.71} \\
     & T2 & 5 & 0.50 & 0.35 & 0.55 & 0.35 & 0.78 & \textbf{0.51} \\
    \cmidrule(lr){2-9}
     & \textbf{Avg} & -- & 0.77 & 0.77 & 0.62 & 0.77 & 0.78 & \textbf{0.74} \\
    \midrule
    \midrule
    \textbf{Theme II} 
    & T3 & 1 & 0.80 & 0.85 & 0.45 & 0.82 & 0.35 & \textbf{0.65} \\
     & T3 & 2 & 0.50 & 0.85 & 0.80 & 0.85 & 0.88 & \textbf{0.78} \\
     & T3 & 3 & 0.60 & 0.85 & 0.65 & 0.78 & 0.90 & \textbf{0.76} \\
    \cmidrule(lr){2-9}
    & T6 & 1 & 0.90 & 0.90 & 0.80 & 0.87 & 0.90 & \textbf{0.87} \\
    & T6 & 2 & 0.50 & 1.00 & 0.65 & 0.92 & 0.75 & \textbf{0.76} \\
    & T6 & 3 & 0.15 & 0.15 & 0.20 & 0.34 & 0.05 & \textbf{0.18} \\
    & T6 & 4 & 0.95 & 0.85 & 0.92 & 0.97 & 0.85 & \textbf{0.91} \\
    \cmidrule(lr){2-9}
    & T7 & 1 & 0.85 & 0.95 & 0.85 & 0.97 & 0.90 & \textbf{0.90} \\
    & T7 & 2 & 0.30 & 0.45 & 0.45 & 0.35 & 0.35 & \textbf{0.38} \\
    & T7 & 3 & 0.85 & 0.95 & 0.85 & 0.97 & 0.95 & \textbf{0.91} \\
    \cmidrule(lr){2-9}
    & T8 & 1 & 0.75 & 0.85 & 0.75 & 0.72 & 0.85 & \textbf{0.78} \\
     & T8 & 2 & 0.75 & 0.30 & 0.38 & 0.25 & 0.25 & \textbf{0.39} \\
     & T8 & 3 & 0.85 & 0.70 & 0.68 & 0.08 & 0.15 & \textbf{0.49} \\
    \cmidrule(lr){2-9}
     & \textbf{Avg} & -- & 0.67 & 0.74 & 0.65 & 0.68 & 0.63 & \textbf{0.67} \\
    \midrule
    \midrule
    \textbf{Theme III} 
    & T4 & 1 & 0.85 & 0.80 & 0.72 & 0.75 & 0.85 & \textbf{0.79} \\
     & T4 & 2 & 0.20 & 0.10 & 0.30 & 0.28 & 0.80 & \textbf{0.34} \\
     & T4 & 3 & 0.90 & 0.90 & 0.65 & 0.87 & 0.90 & \textbf{0.84} \\
     & T4 & 4 & 0.75 & 0.85 & 0.75 & 0.75 & 0.85 & \textbf{0.79} \\
     & T4 & 5 & 0.85 & 0.95 & 0.90 & 0.97 & 0.90 & \textbf{0.91} \\
    \cmidrule(lr){2-9}
    & T5 & 1 & 0.75 & 0.90 & 0.60 & 0.65 & 0.85 & \textbf{0.75} \\
    & T5 & 2 & 0.85 & 0.95 & 0.80 & 0.88 & 0.90 & \textbf{0.88} \\
    & T5 & 3 & 0.85 & 0.90 & 0.65 & 0.90 & 0.88 & \textbf{0.84} \\
    & T5 & 4 & 0.90 & 0.90 & 0.85 & 0.97 & 0.95 & \textbf{0.91} \\
    \cmidrule(lr){2-9}
     & \textbf{Avg} & -- & 0.77 & 0.81 & 0.69 & 0.78 & 0.88 & \textbf{0.79} \\
    \midrule
    \midrule
    \textbf{Theme IV} 
    & T9 & 1 & 0.75 & 0.85 & 0.70 & 0.78 & 0.85 & \textbf{0.79} \\
    & T9 & 2 & 0.92 & 0.90 & 0.70 & 0.87 & 0.90 & \textbf{0.86} \\
    & T9 & 3 & 0.65 & 0.70 & 0.65 & 0.35 & 0.85 & \textbf{0.64} \\
    \cmidrule(lr){2-9}
     & T10 & 1 & 0.85 & 0.70 & 0.70 & 0.80 & 0.90 & \textbf{0.79} \\
     & T10 & 2 & 0.80 & 0.70 & 0.35 & 0.63 & 0.25 & \textbf{0.55} \\
     & T10 & 3 & 0.25 & 0.10 & 0.18 & 0.25 & 0.85 & \textbf{0.33} \\
     & T10 & 4 & 0.75 & 0.75 & 0.60 & 0.88 & 0.85 & \textbf{0.77} \\
    \cmidrule(lr){2-9}
     & \textbf{Avg} & -- & 0.71 & 0.67 & 0.55 & 0.65 & 0.78 & \textbf{0.67} \\
    \midrule
    \textbf{Overall} & -- & -- & \textbf{0.73} & \textbf{0.75} & \textbf{0.63} & \textbf{0.73} & \textbf{0.75} & \textbf{0.72} \\
    \bottomrule
    \end{tabular}%
    }
    \end{table}